%% file: acl_latex.tex
\pdfoutput=1

\documentclass[11pt]{article}

\usepackage[final]{acl}

\usepackage{times}
\usepackage{latexsym}

\usepackage[T1]{fontenc}

\usepackage[utf8]{inputenc}

\usepackage{microtype}

\usepackage{inconsolata}

\usepackage{multirow}
\usepackage{booktabs}

\usepackage{array}
\usepackage{enumitem}
\usepackage{xcolor}
\usepackage{colortbl} 
\usepackage{amsmath}
\usepackage{amsfonts}
\usepackage{hyperref}
\usepackage{graphicx}
\usepackage{makecell}
\usepackage{ragged2e} 
\usepackage{array} 
\usepackage{tabularx}
\usepackage{arydshln}
\usepackage{dashrule}
\usepackage{xspace}
\usepackage{soul}
\usepackage{pgfplots}
\pgfplotsset{compat=1.18}

\newcommand{\Ours}{\textsc{HierPref}\xspace}
\newcommand{\mquakeAdapted}{InstructMH-3k\xspace}
\newcommand{\realCounterMemoryQA}{CounterMemoryMRQA\xspace}

\definecolor{lightblue}{rgb}{0.9,0.9,1}
\sethlcolor{lightblue}

\definecolor{lightred}{rgb}{1,0.8,0.8}
\newcommand{\hlred}[1]{{\sethlcolor{lightred}\hl{#1}}}

\usepackage{tikz}
\usetikzlibrary{patterns}
\definecolor{battleshipgrey}{rgb}{0.3, 0.3, 0.3}
\definecolor{brilliantrose}{rgb}{1.0, 0.33, 0.64}
\definecolor{americanrose}{rgb}{1.0, 0.01, 0.24}
\definecolor{jweigreen}{rgb}{0,0.45,0.24}
\definecolor{bluegray}{rgb}{0.1, 0.1, 0.4}
\definecolor{ao(english)}{rgb}{0.0, 0.5, 0.0}
\definecolor{blanchedalmond}{rgb}{1.0, 0.92, 0.8}
\definecolor{atomictangerine}{rgb}{1.0, 0.6, 0.4}
\definecolor{chocolate(web)}{rgb}{0.82, 0.41, 0.12}
\definecolor{bananayellow}{rgb}{1.0, 0.88, 0.21}
\definecolor{goldenbrown}{rgb}{0.6, 0.4, 0.08}
\definecolor{aliceblue}{rgb}{0.94, 0.97, 1.0}
\definecolor{beige}{rgb}{0.96, 0.96, 0.86}
\definecolor{babyblue}{rgb}{0.54, 0.81, 0.94}
\definecolor{camel}{rgb}{0.76, 0.6, 0.42}
\definecolor{cinnamon}{rgb}{0.82, 0.41, 0.12}

\usepackage{pifont} 
\newcommand{\okmark}{{\textbf{\textcolor[rgb]{0.1, 0.5, 0.1}{$\checkmark$}}}}
\newcommand{\ngmark}{{\textbf{\color{red}{\ding{55}}}}}

\usepackage{tcolorbox}
\NewTColorBox{PromptTemplate}{ O{!htbp} m }{%
    floatplacement={#1},
    float,
    width=\linewidth,
    colback=gray!5!white,
    colframe=gray!75!black,
    title={#2},
}

\newcolumntype{Y}{>{\centering\arraybackslash}X}

\title{Establishing Knowledge Preference in Language Models}

\author{
Sizhe Zhou\textsuperscript{1}, Sha Li\textsuperscript{1}, Yu Meng\textsuperscript{2}, Yizhu Jiao\textsuperscript{1}, Heng Ji\textsuperscript{1}, Jiawei Han\textsuperscript{1} \\
\textsuperscript{1} University of Illinois Urbana-Champaign \quad \textsuperscript{2} University of Virginia  \\
\texttt{\{sizhez, shal2, yizhuj2, hengji, hanj\}@illinois.edu, yumeng5@virginia.edu}
}

\begin{document}
\maketitle

\input{Contents/abstract}

\input{Contents/introduction}

\input{Contents/formulation_of_knowledge_preference}

\input{Contents/evaluating_hierarchical_knowledge_preference}

\input{Contents/methodology}

\input{Contents/tables/tab_ifqa_full-split_main}
\input{Contents/tables/tab_knowledge_preference_3-shot_main}

\input{Contents/experiments}

\input{Contents/tables/tab_mrqa_test}
\input{Contents/tables/tab_real_counter_memory_QA}
\input{Contents/tables/tab_real_counter_memory_QA_data_statistics}
\input{Contents/tables/tab_ablation}

\input{Contents/results_and_analysis}

\input{Contents/related_work}

\input{Contents/conclusion}

\clearpage

\input{Contents/limitations}

\input{Contents/ethics_statements}

\input{Contents/acknowledgements}

\bibliography{custom}

\clearpage
\appendix
\input{Contents/appendix}

\end{document}

%% file: Contents/abstract.tex
\begin{abstract}
Language models are known to encode a great amount of factual knowledge through pretraining. However, such knowledge might be insufficient to cater to user requests, requiring the model to integrate external knowledge sources and adhere to user-provided specifications. 
When answering questions about ongoing events, the model should use recent news articles to update its response; when asked to provide recommendations, the model should prioritize user specifications over retrieved product reviews; when some facts are edited in the model, the updated facts should override all prior knowledge learned by the model even if they are conflicting. 
In all of the cases above, the model faces a decision between its own parametric knowledge, (retrieved) contextual knowledge, and user instruction knowledge. 
In this paper, we (1) unify such settings into the problem of \textit{knowledge preference} and define a three-level preference hierarchy over these knowledge sources; (2) compile a collection of existing datasets IfQA, MQuAKE, and MRQA covering a combination of settings (with/without user specifications, with/without context documents) to systematically evaluate how well models obey the intended knowledge preference; and (3) propose a dataset synthesis method that composes diverse question-answer pairs with user assumptions and related context to directly fine-tune LMs for instilling the hierarchy of knowledge.
We demonstrate that a 7B model, fine-tuned on only a few thousand examples automatically generated by our proposed method, effectively achieves superior performance (more than 18\% improvement across all evaluation benchmarks) in adhering to the desired knowledge preference hierarchy.
\end{abstract}

%% file: Contents/introduction.tex
\begin{figure}[!ht]
    \centering
    \includegraphics[width=\linewidth]{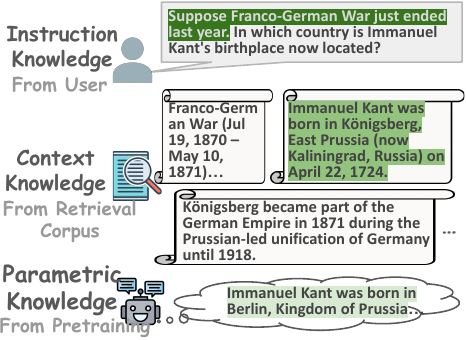}
    \caption{Examples of Instruction Knowledge, Context Knowledge and Parametric Knowledge. The conflicted parts are highlighted. The conflict between the instruction knowledge and the context knowledge lies in the conflicted timestamps. The conflict between the context knowledge and the parametric knowledge lies in the conflicted factual knowledge. }
    \label{fig:motivation}
    \vspace{-15pt}
\end{figure}

\section{Introduction}

Language models memorize factual knowledge during pretraining, which allows them to perform open-domain question answering with remarkable accuracy. 
However, the knowledge encoded within the model (parametric knowledge) might be erroneous or incomplete, falling short of users' expectations. 
Some applications require the language model to leverage the most recent knowledge, such as the latest election results, or stock prices. This is typically set up as closed-domain QA or retrieval-augmented generation (RAG) where the newer knowledge is presented as extra context to the language model. While much effort has been spent on improving retrieval and ranking results, it would be futile if the model simply disregards the input and sticks to its own ``prior beliefs''~\cite{longpre-etal-2021-entity, yu2023information}. Even if the model only occasionally appears obstinate, this will largely undermine user trust as now users would need to fact-check every claim against the provided context. 
In these applications, it is critical to ensure that contextual knowledge is preferred over the models' parametric knowledge. 
Another type of application including personalized search and recommendation requires the integration of user preferences. 
User preferences should always be respected over model parametric knowledge and contextual knowledge. 
Model editing~\cite{meng2022locating, meng2022mass, de2021editing, mitchell2022memory, zhong2023mquake}  can be seen as a special case of such preferences, where the new facts override learned facts even if they are counterfactual in nature. 
In all of these settings (RAG, closed-domain QA, integrating user beliefs and model editing), we observe that the key is to enforce a certain priority among knowledge from different sources. 

The strife between parametric knowledge and contextual knowledge has been measured across many models and forms of contexts~\cite{longpre-etal-2021-entity, neeman-etal-2023-disentqa,li-etal-2023-large, xie2023adaptive, kortukov2024studying}. While earlier models (T5~\cite{raffel2020exploring}, Roberta~\cite{liu2019roberta}) seem to be baffled by conflicting knowledge and often stick to their priors~\cite{longpre-etal-2021-entity}, recent larger models (OPT~\cite{zhang2022opt}, GPT-3~\cite{brown2020language}) show potential in successfully updating their answers through in-context edits~\cite{zheng-etal-2023-edit, zhong2023mquake, si2023promptinggpt3reliable, kortukov2024studying}. Existing studies also reveal some influence factors for in-context update failures, such as incoherence context~\cite{xie2023adaptive} and parametric answers (the answer according to parametric knowledge) appearing in context~\cite{kortukov2024studying}.
Under the RAG setting, attempts have been made to rectify model behavior in the presence of noisy retrieval~\cite{zhang2024raftadaptinglanguagemodel, yoran2024makingretrievalaugmentedlanguagemodels}, requiring the model to cite retrieved contextual knowledge only when it is relevant to the question.
While these lines of work are seemingly separate, we believe that they are just shapes and forms of the same underlying question: \textit{how should language models behave when faced with multiple sources of (noisy) knowledge?} 

To answer this question, we first build our framework of hierarchical knowledge preference over three distinct levels: parametric knowledge, contextual knowledge and instruction knowledge. While the divide between parametric and contextual knowledge is not new, we make the further distinction between (retrieved) contextual knowledge and (user or system-provided) instruction knowledge to account for the case of noisy context.
This three-level hierarchy unifies multiple settings: (1) prioritizing instruction knowledge over parametric knowledge is the problem of in-context knowledge editing~\cite{zheng-etal-2023-edit}; (2) prioritizing contextual knowledge over parametric knowledge is the problem of RAG and closed-domain QA~\cite{zhang2024raftadaptinglanguagemodel, yoran2024makingretrievalaugmentedlanguagemodels}; (3) the full hierarchy supports personalized or counterfactual QA with RAG~\cite{yu-etal-2023-ifqa}.

To systematically evaluate a model's ability to adhere to the desired knowledge preference hierarchy, we create a benchmark adapted from several existing datasets (IfQA~\cite{yu-etal-2023-ifqa}, MQuAKE~\cite{zhong2023mquake} and MRQA~\cite{fisch2019mrqa}) to cover all of the aforementioned settings. 
Moreover, we stress-test the model's behavior in more difficult cases where the contextual knowledge is noisy and the question requires reasoning (multi-hop). We observe that while large, proprietary models such as GPT-4o can perform relatively well (86.46\% F$_1$ on the counterfactual knowledge editing task), open-source models, especially those fine-tuned with open instruction data (Mistral with Alpaca tuning only achieves 28.48\% F$_1$ on same task), fail to model this knowledge hierarchy even when they are explicitly instructed to do so in the prompt. 

To close this gap, we design a dataset synthesis procedure to create instruction-tuning data that follows our desired order of knowledge preference. 
We start from Wikipedia and Wikidata, which are known as high-quality sources of factual data, and use GPT-4o to synthesize questions and counterfactual evidence. 
For multi-hop questions, we sample fact chains from Wikidata, alter some of the intermediate facts, and then synthesize passages to support each hop. 
Our dataset creation process does not rely on any human annotation and through experiments, we show that a few thousand examples are sufficient to unlock the knowledge preference ability of open-source LLMs (28.48\% F$_1$ $\rightarrow$ 89.36\%  F$_1$ on the counterfactual knowledge editing task without specific prompting). Our model is also more robust when encountering noisy knowledge and shows even more gains on complex, multi-hop questions. 

To conclude, our main contributions include: 
\begin{itemize}[leftmargin=*,nosep]
    \item We formulate the \textit{knowledge preference} problem of LLMs, which unifies many settings where the model needs to decide among parametric knowledge, contextual knowledge, and user instruction knowledge. 
    \item We compile a benchmark to evaluate the knowledge preference property of LLMs by adapting existing datasets to cover all combinations of different settings and difficulties. We encourage model developers to take \textit{knowledge preference} as an additional axis of evaluation as many important applications (RAG, knowledge editing, and user preference modeling) entail this ability. 
    \item We design a data synthesis procedure to automatically create instruction-tuning data for instilling the knowledge preference. 
    We show that fine-tuning an open-source language model with a few thousand dedicated data samples can make the model much more receptive to user instruction knowledge and contextual knowledge, achieving superior performance on all settings in our benchmark.
\end{itemize}

%% file: Contents/formulation_of_knowledge_preference.tex
\section{Formulation of Knowledge Preference}
When the parametric knowledge (intrinsic knowledge)~\cite{petroni2019language, mallen2022not} of an LLM is insufficient to give the correct answer to user queries, we can introduce external knowledge either in the instruction or as additional context.

\paragraph{Instruction Knowledge} is the knowledge injected through user instructions.  
Instruction knowledge can refer to rules or principles that govern how the model should utilize other types of knowledge, i.e. problem-solving constraints from user instructions and assumptions from hypothetical questions.

\paragraph{Context Knowledge} is the potentially noisy context provided to the LLM during inference time. 
One typical case is the retrieved passages in retrieval-augmented generation. 
The retrieved passages can provide newly-updated knowledge or domain-specific knowledge which is generally expected to override or complement LLMs' own knowledge in RAG. 

We take the RAG case in Fig.~\ref{fig:motivation} as an example where the user queries the LLM with a question (ignore the question assumption first). 
Resolving the question requires solving a model preference problem where we want the LLM to prioritize relevant knowledge in the retrieved context over knowledge embedded in the LLM's parameters. 
Sometimes, users will give their own constraints or requirements for answering the query (e.g., the question assumption in Fig.~\ref{fig:motivation}).
Correspondingly, to fulfill the user requirements, the LLM should override the original way it utilizes the knowledge, by following a new reasoning flow and utilizing different pieces of context knowledge and parametric knowledge. 
Then, the RAG case in Fig.~\ref{fig:motivation} is fundamentally a knowledge preference problem where we further give the instruction knowledge the highest priority in the inference process.

More generally, in this work, 
we define the principle of \emph{Hierarchical Knowledge Preference} built on these three types of knowledge. 

\paragraph{Hierarchical Knowledge Preference} 
In applications of LLMs, conflicts between instruction knowledge, context knowledge, and parametric knowledge are frequently inevitable.
For instance, a user may provide counterfactual hypothesis or unprecedented constraints which may conflict with the retrieved documents or the LLMs' own knowledge~\cite{yu-etal-2023-ifqa}. 
Meanwhile, the retrieved documents serving as the context knowledge may bring facts which disagree with LLMs' outdated or wrong memory~\cite{vu2023freshllms}.  
Ignorance or inappropriate handling of these knowledge conflicts can result in nondeterministic inference behaviors of LLMs, thus undermining downstream LLM-based applications.

We define our hierarchy of ideal knowledge preference as follows:

(i) \emph{Instruction Knowledge $\succ$ Context Knowledge}. 
The knowledge from the instruction should be accorded the highest priority so that LLMs can orient all of the reasoning power or acquired knowledge toward fulfilling the system-level or user-level requirements.

(ii) \emph{Context Knowledge $\succ$ Parametric Knowledge}. 
As the parametric knowledge is mainly acquired in the pre-training stage which restricts the parametric knowledge itself to be timely corrected, updated, or expanded, we assume the retrieved or given context knowledge should be generally preferred at the time of inference.\footnote{In the knowledge conflict scenarios where the context or the retrieved contents are flawed (e.g., misleading or not completely accurate), models' own parametric knowledge could be more reliable. In this work, we assume the retrieved contents are generally helpful and should be prioritized over parametric knowledge. Otherwise there is no such need for RAG in such scenarios. }
Note that our knowledge preference is defined for the scenarios where direct knowledge conflicts arise.
This means that the information irrelevant to solving the target problem or answering the target query should be regarded as noise and it does not contribute to any knowledge conflicts.

%% file: Contents/evaluating_hierarchical_knowledge_preference.tex
\section{Benchmark Construction}\label{sec::Evaluating Hierarchical Knowledge Preference}
As prior works mainly focus on the conflicts between external context knowledge and the parametric knowledge~\cite{xie2023adaptive} or conflicts within a single type of knowledge~\cite{wallace2024instruction}, there is a lack of a comprehensive and high quality evaluation benchmark for evaluating hierarchical knowledge preference. 

\subsection{Evaluating Preference for Instruction Knowledge}
To evaluate LLMs' preference for instruction knowledge, we focus on the case where counterfactual assumptions are introduced by the instruction, which is a typical scenario calling for the preference for instruction knowledge and it's more likely to introduce explicit and direct knowledge conflicts between the instruction knowledge and other types of knowledge.

Among existing works, IfQA~\cite{yu-etal-2023-ifqa} is a human annotated counterfactual QA benchmark where the question introduces hypothetical conditions. 
We adopt the test set of its full split which has 700 instances in total for evaluating the priority of instruction knowledge in retrieval-augmented setting. 
We utilize two setups for retrieval augmented setting: (i) \texttt{w/ Gold Passages} where the oracle context following the question is given, and (ii) \texttt{w/ Mixed Passages} where the top-3 retrieved passages from Wikipedia dump along with the oracle contexts and the question is given to be more realistic. 
The F$_1$ and Exact Matching (EM) scores are reported.

However, the knowledge conflicts introduced by IfQA may not be explicit and significant enough. 
For example, in the question \textit{If sea levels had risen significantly over the past decade, which country would have been the first to be submerged?}, the instruction knowledge \textit{sea levels had risen significantly over the past decade} does not directly conflict with the oracle context passage which is about \textit{the world's lowest-lying country}.

Therefore, we further extend a knowledge editing benchmark MQuAKE-CF-3k~\cite{zhong2023mquake} to be \mquakeAdapted  for evaluating testee LLMs' preference between instruction knowledge and the context knowledge.
MQuAKE-CF-3k contains multi-hop QA instances based on human-filtered relations, entities, and crafted templates for verbalizing relation triples, but without context passages. 
Each relation triple is guaranteed to be recallable by GPT-J~\cite{gpt-j}.
Each multi-hop QA instance is associated with a fact chain (sequentially linked relation triples), and knowledge edits. 
So we integrate the knowledge edits with the original question to obtain a counterfactual multi-hop question (see the question in Fig.~\ref{fig:case_study_instruction_knowledge} for an example). 
For each factual relation triple needed to get to both the original answer before fact chain editing and the new answer after fact chain editing, we adopt GPT-3.5 to synthesize one supporting context passage which will be given along with the question to the testee LLMs.
We evaluate the F$_1$ and EM scores according to both the original answer and the new answer.
If testee LLMs well prioritize the instruction knowledge and generally prefer context knowledge than parametric knowledge, they should follow the counterfactual instruction assumptions, focus on the suitable passages in the context, and reach the new answer instead of the original answer, leading to a higher evaluation scores with new answers than with original answers.

\subsection{Evaluating Preference for Context Knowledge}
To evaluate LLMs' preference for context knowledge, we adopt the test set of MRQA~\cite{fisch2019mrqa}, covering BioASQ~\cite{tsatsaronis2015overview}, DROP~\cite{dua2019drop}, DuoRC~\cite{saha2018duorc}, RACE~\cite{lai2017race}, RelationExtraction~\cite{levy2017zero}, and TextbookQA~\cite{kembhavi2017you} across various domains.
We divide the evaluation into two parts. 
The first part is the evaluation on the open-book QA on the whole test set, denoted as MRQA. 
This quantifies the generally capability of testee LLM to comprehend and prioritize the context knowledge regardless of whether the context knowledge conflicts with their parametric knowledge or not. 
Here F$_1$ and EM are reported.

The second part of the evaluation (denoted as \realCounterMemoryQA) is conducted on the subset of the test set where LLMs' parametric knowledge is conflicted with the context knowledge. 
So we first probe the parametric knowledge of each testee LLM with 3-shot exemplars (taken from MRQA dev set) to obtain the target test subset. 
Then, we measure the proportion of test instances for which the model correctly updates its answer (denoted as $\mathbb{P}(\mathbf{U_c})$) and the proportion of test instances for which model incorrectly updates its answer (denoted as $\mathbb{P}(\mathbf{U_i})$).\footnote{We decide whether an answer is the same as the gold-standard answers or not, we use F$_1$ to tolerate minor deviates and set F$_1$ higher than 0.8 as the same and F$_1$ lower than 0.2 as different.} 
If a testee LLM well prioritizes the context knowledge, $\mathbb{P}(\mathbf{U_c})$ should be significantly higher than $\mathbb{P}(\mathbf{U_i})$.

%% file: Contents/methodology.tex
\section{Methodology}

In this work, compared to designing prompt strategies to constrain LLMs with the hierarchical knowledge preference, we choose to inherently embed the hierarchical knowledge preference inside LLMs which is versatile and potentially benefits broader tasks.
Hence, we resort to instruction tuning which is shown effective in aligning LLMs' behaviors with human expectations~\cite{wei2021finetuned}. 
We model the hierarchical knowledge preference behavior of LLMs through the synthesis of corresponding instruction tuning data. 

First, we acquire diverse and high-quality passages and fact chains from Wikipedia and Wikidata\footnote{\href{https://www.wikidata.org/wiki/Wikidata:Main_Page}{WikiData main page}} as source data for subsequent synthesis (Sec.~\ref{sec::Methodology::Source Data Collection}).
The target types of our synthesized data are designed to include both single-hop and multi-hop QA. 
Second, we teach LLMs to prioritize instruction knowledge through synthesizing counterfactual retrieval-augmented QA data (Sec.~\ref{sec::Methodology::Modeling Preference for Instruction Knowledge}).
Third, we also teach LLMs to prioritize context knowledge over their parametric knowledge by synthesizing factual retrieval-augmented QA data with context-supported answer conflicting with LLMs' parametric answer (Sec.~\ref{sec::Methodology::Modeling Preference for Context Knowledge}).
Final statistics of synthesized data can be seen in Appendix~\ref{sec::Appendix::Statistics of Synthesized Data}.

\subsection{Source Data Collection}\label{sec::Methodology::Source Data Collection}
\begin{figure}[!ht]
    \centering
    \includegraphics[width=0.92\linewidth]{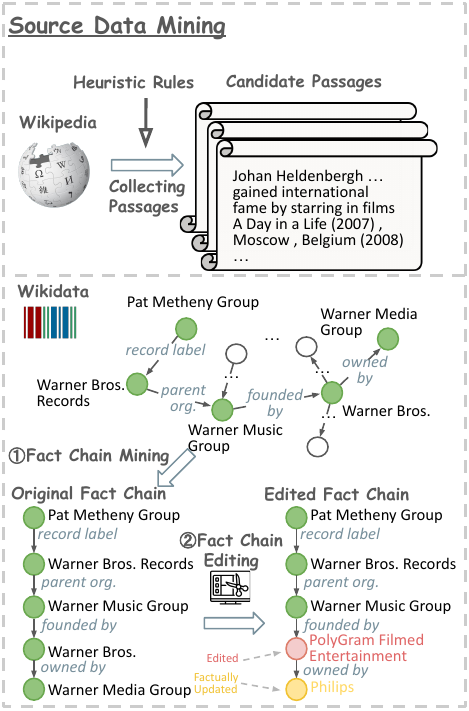}
    \caption{
    \texttt{Source Data Collection} step of \Ours synthesis framework.
    }
    \label{fig:framework_step1}
    \vspace{-12pt}
\end{figure}

\begin{figure*}[!t]
    \centering
    \includegraphics[width=\linewidth]{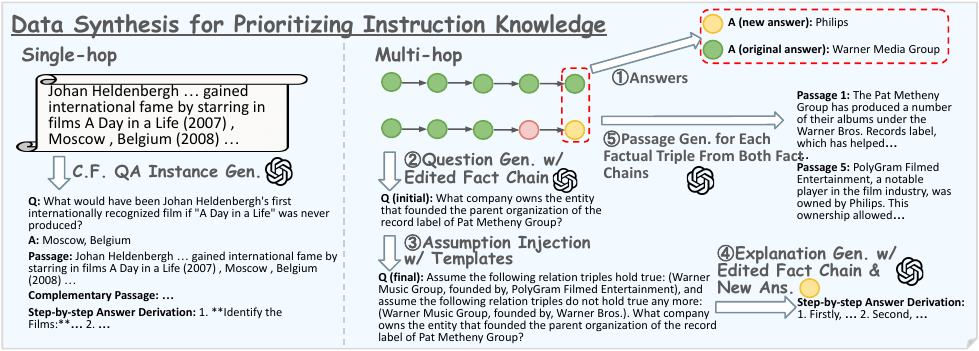}
    \caption{
    \texttt{Modeling Preference for Instruction Knowledge} step of \Ours synthesis framework.
    \texttt{C.F.} denotes \texttt{Counter Factual}.
    }
    \label{fig:framework_step2}
    \vspace{-12pt}
\end{figure*}

In terms of the instance contents, in contrast to synthesis-based approaches which rely on LLMs to synthesize the entire input and output of each instance, our goal is to provide maximal control on the synthesized contents while ensuring the expected quality.
In terms of the data format, we mainly focus on the single-hop and the multi-hop question answering data given reference passages which is related to broad downstream applications of LLMs, especially in the retrieval-augmented setting.

First, we gather a corpus of Wikipedia passage chunks as oracle contexts for subsequent single-hop QA data synthesis. 
To enhance the efficiency of the corpus to serve for fact-related QA data synthesis, we trace back to the Wikipedia passages that contain evidence for verifiable instances from the FEVER dataset~\cite{thorne2018fever}.
A heuristic rule is further applied to filter passages whose number of distinct named entities are fewer than 5. 
This step results in a corpus of high-quality Wikipedia passages denoted as $\mathcal{C}$.

Second, we traverse the Wikidata to extract a set of fact chains ranging from 2 hops to 4 hops\footnote{We assume the questions with the number of hops exceeding 4 are relatively rare in reality.} for multi-hop QA data synthesis. 
The underlying traversal algorithm is based on breadth-first search (BFS) on the knowledge graph.
Our fact chain mining algorithms targets at mining both a fact chain $l_i$ and its counterfactually edited derivative $l_i'$.
Suppose each fact chain $l_i$ with $m_i$ hops acquired from BFS is $\displaystyle [e^i_0, r^i_0, e^i_1, r^i_1, \dots, r^i_{m_i-1}, e^i_{m_i}]$ which consists of triples $\displaystyle (e^i_0, r^i_0, e^i_1), \dots, (e^i_{m_i-1}, r^i_{m_i-1}, e^i_{m_i})$ in order.
We will randomly choose the number of edits applied on $l_i$ as $K_i$ and recursively conduct the edit one by one. 
Each edit is conducted over the previously edited fact chain. 
At each edit, we will first randomly choose one relation triple from the fact chain (while allowing enough subsequent relation triples for remaining edits) and replace the tail entity with an counterfactual entity of the same type, similar to the misinformation training data generation approach proposed by~\cite{Fung2021}.
Then all the relation triples after this edited relation triple will update their entities factually following this newly changed tail entity without changing any relation.
This completes one edit on the fact chain, resulting in a different fact chain. 
Completing all the $K_i$ edits eventually leads to $l_i'$ as the counterfactually edited derivative of $l_i$.

Please refer to Appendix~\ref{sec::Appendix::Fact Chain Mining} for more details including the heuristic rules applied for the diversity and quality of the mined fact chains. 
The set of candidate original and edited fact chains extracted in this step is denoted as $\mathcal{F}$.

For data synthesis in Sec.~\ref{sec::Methodology::Modeling Preference for Instruction Knowledge} and Sec.~\ref{sec::Methodology::Modeling Preference for Context Knowledge}, we randomly sample a set of Wikipedia passages $\displaystyle \{d_i\}_{i=1}^n \subset \mathcal{C}$ and a set of original and edited Wikidata fact chains $\displaystyle \{(l_i, l_i')\}_{i=1}^m \subset \mathcal{F}$ respectively for each step.

\begin{figure*}[!t]
    \centering
    \includegraphics[width=\linewidth]{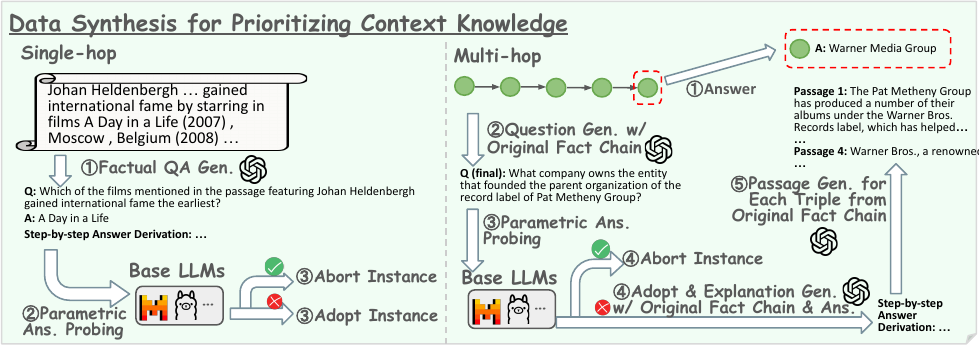}
    \caption{
    \texttt{Modeling Preference for Context Knowledge} step of \Ours synthesis framework.
    \texttt{Data Synthesis for Prioritizing Instruction Knowledge} of Fig.~\ref{fig:framework_step2} and \texttt{Data Synthesis for Prioritizing Context Knowledge} here share the same example source data in Fig.~\ref{fig:framework_step1}. In implementation, two stages' source data have no overlap. 
    }
    \label{fig:framework_step3}
    \vspace{-12pt}
\end{figure*}

\subsection{Modeling Preference for Instruction Knowledge}\label{sec::Methodology::Modeling Preference for Instruction Knowledge}

To synthesize instruction tuning data which grants the highest preference priority for instruction knowledge, we resort to counterfactual question answering. 
The counterfactual assumptions or hypotheses set up the instruction knowledge which will directly conflict with the parts of the factual ``retrieved passages'' and likely deviates from the LLMs' parametric knowledge. 
Such synthesized data can guide LLMs to prioritize the instruction knowledge, overriding conflicted parts of the context knowledge and potentially the parametric knowledge, to reach the correct answer. 

Specifically, for each randomly sampled passage $d_i$, we prompt GPT-4o based on $d_i$ to synthesize an single-hop QA instance containing: 
\begin{itemize}[leftmargin=*,nosep]
    \item The counterfactual question which introduces counterfactual and hypothetical conditions or incidents.
    \item The precise, concise, no-trivial, and uniquely-derivable answer through counterfactual reasoning based on $d_i$, the hypothetical question, and common sense\footnote{As counterfactual reasoning might inherently use some common sense knowledge beyond the context and the question, and it's hard to elaborate them one by one, we do not prevent GPT-4o from using them.}.
    \item Extra information as an additional passage to make sure the answer is uniquely derivable.
    \item The step-by-step answer derivation explanation. 
\end{itemize}
Please refer to Appendix~\ref{sec::Appendix::Prompt Templates::Data Synthesis Prompt Templates} for prompt templates used to obtain these components.
Through prompting GPT-4o for instance synthesis, we expect that GPT-4o can bring more diversity and non-trivial difficulty through leveraging its reasoning power and external knowledge beyond the provided Wikipedia passage $d_i$. 
Human annotators could provide higher quality for this kind of data as they can be better at recalling related external knowledge and capturing their underlying associations through complex reasoning. 
However, the disadvantages of relying human annotators include the expense and the potentially limited counterfactual reasoning patterns that human annotators can think of.
To encourage diversity, we adopt no in-context demonstrations for synthesis.

The synthesis for multi-hop QA instance is similar except that the counterfactual assumption is predefined by the counterfactual fact chain edits and the target answer is just the tail entity of the edited fact chain. 
We mainly prompt GPT-4o for synthesizing based on $(l_i, l_i')$:
\begin{itemize}[leftmargin=*,nosep]
    \item The multi-hop question that starts from and includes only the head entity of the edited fact chain $l_i'$, incorporates all the relations, and has the tail entity of $l_i'$ as the final answer. Later we will apply a template to integrate the counterfactual edits as the assumptions with the generated multi-hop question. 
    \item A list of passages for all factual relation triples from $l_i$ and $l_i'$ so that each factual relation triple can be uniquely derived given all of the passages simultaneously. 
    \item The step-by-step answer derivation explanation. 
\end{itemize}
Please refer to Appendix~\ref{sec::Appendix::Prompt Templates::Data Synthesis Prompt Templates} for prompt templates used to obtain these components.
Since we can only mine relation triples from Wikidata, we adopt GPT-4o for synthesis relying on its power to understand and verbalize the relation triples into fluent and coherent natural language. 
To ensure the quality of synthesized multi-hop questions, we took a fixed set of 5 exemplars demonstrating the synthesis of multi-hop question from a given fact chain.

\subsection{Modeling Preference for Context Knowledge}\label{sec::Methodology::Modeling Preference for Context Knowledge}

The goal of modeling the preference for context knowledge is to teach LLMs to prefer the ``retrieved contexts'' over their own parametric knowledge. 
Sticking to the data format of single-hop and multi-hop QA with reference passages, we achieve this goal by synthesizing factual QA instances with answers supported by the passages but opposed by the LLMs' parametric knowledge. 

For single-hop QA instances, we prompt GPT-4o with passage $d_i$ to synthesize the factual question, the corresponding answer, the step-by-step answer derivation, and an additional passage to further make sure the answer is uniquely derivable from the contexts. 
For multi-hop QA instances, we leverage the unedited fact chain $l_i$ and prompt GPT-4o to synthesize the multi-hop question, a list of passages verbalized from relation triples of $l_i$ to ensure the tail entity of $l_i$ is uniquely derivable, and the step-by-step answer derivation.
One special design is that, we will first probe a list of base LLMs with the synthesized question to filter questions that can be correctly answered by the base LLMs' parametric knowledge. 
This step is done before further synthesizing the remaining components of the new instance for efficiency. 
Please refer to Appendix~\ref{sec::Appendix::Prompt Templates::Data Synthesis Prompt Templates} for prompt templates used here.

%% file: Contents/tables/tab_ifqa_full-split_main.tex
\begingroup
\begin{table*}[!ht]
    \small 
    \centering
    \renewcommand\tabcolsep{4pt} 
    \scalebox{1}{
    \begin{tabular}{lccccccccc}
        \toprule 
        \multirow{3}{*}{Model} & \multirow{3}{*}{\# Shots} &  \multicolumn{4}{c}{Normal Prompt} & \multicolumn{4}{c}{Explicit Prompt} \\ 
         \cmidrule(l){3-6} \cmidrule(l){7-10}
                               &                              & \multicolumn{2}{c}{w/ Gold Passages} & \multicolumn{2}{c}{w/ Mixed Passages} & \multicolumn{2}{c}{w/ Gold Passages} & \multicolumn{2}{c}{w/ Mixed Passages} \\
        \cmidrule(l){3-4} \cmidrule(l){5-6} \cmidrule(l){7-8} \cmidrule(l){9-10}
                               &                              & F$_1$ & EM & F$_1$ & EM  & F$_1$ & EM & F$_1$ & EM \\ 
        \midrule
        \rowcolor[gray]{0.85}
        \multicolumn{10}{c}{\textit{\textbf{Reference Models}}}   \\ 
        \midrule
        \multirow{1}{*}{GPT-3.5 Turbo} & 5 & 77.70 & 71.86 & 73.27 & 67.57 & 79.70 & 74.14 & 72.24 & 66.57 \\
        \multirow{1}{*}{GPT-4o} & 0 & 88.09 & 80.43 & 85.39 & 77.86 & 88.19 & 80.71 & 85.38 & 77.29 \\
                                & 3 & 89.56 & 83.29 & 87.12 & 80.71 & 90.18 & 84.43 & 87.87 & 81.29 \\
                                & 5 & 90.43 & 84.57 & 87.50 & 81.14 & 89.71 & 83.86 & 87.88 & 81.57 \\ 
        \midrule
        \rowcolor[gray]{0.85}
        \multicolumn{10}{c}{\textit{\textbf{Main Models}}}   \\ 
        \midrule
        \multirow{1}{*}{Mistral-v0.3-7B} & 3 & 59.52 & 52.14 & 42.34 & 36.43 & 59.56 & 53.43 & 40.27 & 35.00 \\
        \multirow{1}{*}{Mistral-v0.3-7B-Instruct} & 5 & 71.26 & 63.14 & 59.13 & 51.71 & 70.76 & 62.29 & 57.03 & 49.71 \\
        \multirow{1}{*}{Mistral-v0.3-7B w/ Alpaca} & 5 & 67.98 & 61.71 & 50.71 & 44.00 & 67.22 & 60.29 & 49.49 & 43.14 \\ 
        \multirow{1}{*}{Mistral-v0.3-7B w/  \Ours} & 0 & 80.53 & 74.14 & 77.85 & 70.86 & 80.53 & 73.86 & 77.33 & 70.29 \\
        \bottomrule
    \end{tabular}
    }
    \caption{Evaluation results (\%) on IfQA full split test set. Zero shot performance of \Ours is presented and best performance of baselines among \{0, 3, 5\} shots are presented. See Table~\ref{tab:eval_ifqa_full_split} for full results. Assumption-in-Question version of the explicit prompting is applied.}
    \label{tab:main_eval_ifqa_full_split}
\end{table*}
\endgroup

%% file: Contents/tables/tab_knowledge_preference_3-shot_main.tex
\begin{table*}[!ht]
    \small 
    \centering
    \begin{tabular}{lcccccccc}
        \toprule 
        \multirow{2}{*}{Model} & \multicolumn{4}{c}{Normal Prompt} & \multicolumn{4}{c}{Explicit Prompt}  \\ 
        
        \cmidrule(l){2-5} \cmidrule(l){6-9}
                               & F$_1$ & F$_1$ Ratio & EM & EM Ratio & F$_1$ & F$_1$ Ratio & EM & EM Ratio \\ 
        \midrule
        \rowcolor[gray]{0.85}
        \multicolumn{9}{c}{\textit{\textbf{Reference Models}}}   \\ 
        \midrule
        \multirow{1}{*}{GPT-3.5 Turbo} & 34.08 & 0.61 & 32.16 & 0.62 & 35.55 & 0.65 & 33.58 & 0.66 \\
        \multirow{1}{*}{GPT-4o} & 86.46 & 7.63 & 85.61 & 8.99 & 93.37 & 19.23 & 92.54 & 30.62 \\
        \midrule
        \rowcolor[gray]{0.85}
        \multicolumn{9}{c}{\textit{\textbf{Main Models}}}   \\ 
        \midrule
        \multirow{1}{*}{Mistral-v0.3-7B} & 48.16 & 1.20 & 46.64 & 1.24 & 48.95 & 1.23 & 47.36 & 1.27 \\
        \multirow{1}{*}{Mistral-v0.3-7B-Instruct} & 33.34 & 0.76 & 31.12 & 0.81 & 33.42 & 0.77 & 31.12 & 0.81 \\
        \multirow{1}{*}{Mistral-v0.3-7B w/ Alpaca} & 28.40 & 0.50 & 26.28 & 0.49 & 28.48 & 0.50 & 26.34 & 0.49 \\
        \multirow{1}{*}{Mistral-v0.3-7B w/ \Ours} & 89.36 & 10.85 & 88.24 & 14.26 & 89.49 & 11.15 & 88.36 & 14.73 \\
        \bottomrule
    \end{tabular}
    \caption{3-shot evaluation results on \mquakeAdapted. F$_1$ and EM scores are reported in \%. For explicit prompting results, we here present the Assumption-in-Question explicit prompt version which gives generally better performance for target baselines. Table~\ref{tab:eval_knowledge_preference_3-shot} contains full results. }
    \label{tab:main_eval_knowledge_preference_3-shot}
    \vspace{-12pt}
\end{table*}

%% file: Contents/experiments.tex
\section{Experiments}
To validate whether our synthesized data can inherently build LLMs' hierarchical knowledge preference, we fine-tune base LLMs with Alpaca's 52K instruction tuning data plus our \textasciitilde 7.4K \Ours data and evaluate the resulting LLMs on benchmarks elaborated in Sec.~\ref{sec::Evaluating Hierarchical Knowledge Preference}.
Please refer to Appendix~\ref{sec::Appendix::Implementation Details} for implementation details.

\subsection{Prompting for Hierarchical Knowledge Preference}
Without tuning the LLMs, we also experimented with different prompts to see whether they can enhance or establish the hierarchical knowledge preference. 
In this work, we mainly apply three prompting templates (see Appendix~\ref{sec::Appendix::Prompt Templates::Strong Prompts for Hierarchical Knowledge Preference}): 
(i) Alpaca~\cite{alpaca}'s prompt template as baseline. 
(ii) \texttt{Assumption-in-Instruction} based on (i) which puts instruction knowledge in the instruction and the instruction explicitly requires LLMs to follow the hierarchical knowledge preference,
(iii) \texttt{Assumption-in-Question} based on (i) which puts instruction knowledge along with the question in the input and the instruction explicitly requires LLMs to follow the hierarchical knowledge preference.
We denote (i) as \texttt{Normal Prompt} and denote (ii) and (iii) as \texttt{Explicit Prompt}.

\subsection{Evaluation Baselines}
Our comparison mainly focuses on the base LLM trained with Alpaca's 52K instruction tuning data (denoted as \texttt{w/ Alpaca}) and the same base LLM trained with the same 52K data plus our \Ours data (denoted as \texttt{w/} \Ours). 
We select Mistral-v0.3-7B released in 05/22/2024 as the base LLM. 
In addition to this, we also include LLMs include Llama-2~\cite{touvron2023llama}, Llama-3~\cite{llama3modelcard},  Qwen-2~\cite{qwen}, GPT-3.5~\cite{openai2023gpt35}, and GPT-4o~\cite{openai2023gpt4o} with both the base model and instruction-tuned model for reference.

%% file: Contents/tables/tab_mrqa_test.tex
\begingroup
\begin{table*}[!ht]
    \small 
    \centering
    \renewcommand\tabcolsep{3pt} 
    \scalebox{0.88}{
    \begin{tabular}{lccccccccccccccc}
        \toprule 
        \multirow{2}{*}{Model} & \multirow{2}{*}{SP} & \multicolumn{2}{c}{Overall} & \multicolumn{2}{c}{BioASQ} & \multicolumn{2}{c}{DROP} & \multicolumn{2}{c}{DuoRC} & \multicolumn{2}{c}{RACE} & \multicolumn{2}{c}{RE} & \multicolumn{2}{c}{TextbookQA} \\ 
        \cmidrule(l){3-4} \cmidrule(l){5-6} \cmidrule(l){7-8} \cmidrule(l){9-10} \cmidrule(l){11-12} \cmidrule(l){13-14} \cmidrule(l){15-16}
                               &                                   & F$_1$ & EM                  & F$_1$ & EM                 & F$_1$ & EM               & F$_1$ & EM                             & 
        F$_1$ & EM               & F$_1$ & EM                             & F$_1$ & EM               \\
        \midrule
        \multirow{2}{*}{Mistral-v0.3-7B w/ Alpaca} & \okmark & 54.94 & 41.27 & 53.24 & 30.92 & 42.32 & 30.01 & 38.80 & 24.65 & 31.35 & 16.47 & 83.45 & 72.90 & 40.02 & 28.61 \\
                                                   & \ngmark  & 56.81 & 42.99 & 55.84 & 32.45 & 44.45 & 32.53 & 40.59 & 26.18 & 33.64 & 18.25 & 84.56 & 74.08 & 42.29 & 30.87 \\
        \multirow{2}{*}{Mistral-v0.3-7B w/ Alpaca 3-shot} & \okmark & 60.51 & 48.29 & 65.19 & 45.21 & 50.70 & 39.25 & 45.00 & 33.64 & 35.90 & 22.26 & 82.78 & 72.39 & 48.47 & 39.45 \\
                                                          & \ngmark  & 60.66 & 48.39 & 65.35 & 45.74 & 51.50 & 39.92 & 44.66 & 32.64 & 39.17 & 25.37 & 82.58 & 72.42 & 47.75 & 38.39 \\
        \multirow{2}{*}{Mistral-v0.3-7B w/ \Ours} & \okmark & 73.52 & 63.01 & 79.31 & 64.10 & 61.66 & 52.69 & 63.28 & 51.03 & 56.96 & 43.47 & 88.75 & 80.63 & 67.39 & 58.42 \\
                                                  & \ngmark  & 73.67 & 62.91 & 79.53 & 63.50 & 61.39 & 52.10 & 63.41 & 51.23 & 57.16 & 43.18 & 88.58 & 80.43 & 68.51 & 59.28 \\
        \bottomrule
    \end{tabular}
    }
    \caption{Evaluation results (\%) on MRQA given oracle contexts. Here \texttt{SP} refers to whether the explicit prompting strategy of Assumption-in-Question is applied or not. }
    \label{tab:eval_mrqa_test_w_oracle_contexts}
\end{table*}
\endgroup

%% file: Contents/tables/tab_real_counter_memory_QA.tex
\begin{table*}[!t]
    \small 
    \centering
    \scalebox{1}{
    \begin{tabular}{lcccccccc}
        \toprule 
        \multirow{3}{*}{Dataset} &  \multicolumn{4}{c}{Normal Prompt}  &  \multicolumn{4}{c}{Explicit Prompt}  \\ 
        \cmidrule(l){2-5} \cmidrule(l){6-9}
                                 &  \multicolumn{2}{c}{Mistral w/ Alpaca} &  \multicolumn{2}{c}{Mistral w/ \Ours} &  \multicolumn{2}{c}{Mistral w/ Alpaca} &  \multicolumn{2}{c}{Mistral w/ \Ours} \\
        \cmidrule(l){2-3} \cmidrule(l){4-5} \cmidrule(l){6-7} \cmidrule(l){8-9}
                                 & $\mathbb{P}(\mathbf{U_i})$ & $\mathbb{P}(\mathbf{U_c})$ & $\mathbb{P}(\mathbf{U_i})$ & $\mathbb{P}(\mathbf{U_c})$ & $\mathbb{P}(\mathbf{U_i})$ & $\mathbb{P}(\mathbf{U_c})$ & $\mathbb{P}(\mathbf{U_i})$ & $\mathbb{P}(\mathbf{U_c})$ \\
        \midrule
        BioASQ & 31.47 & 41.45 & 16.89 & 61.15 & 31.47 & 41.30 & 17.54 & 61.64 \\
        DROP & 46.50 & 35.09 & 40.33 & 45.83 & 47.17 & 34.23 & 39.74 & 46.91 \\
        DuoRC.ParaphraseRC & 48.74 & 31.48 & 31.56 & 50.00 & 49.04 & 32.44 & 31.63 & 49.63 \\
        RACE & 54.62 & 23.56 & 36.38 & 42.53 & 56.37 & 20.42 & 36.20 & 42.36 \\
        RelationExtraction & 13.11 & 70.19 & 8.73 & 78.29 & 12.26 & 70.35 & 8.56 & 78.40 \\
        TextbookQA & 61.88 & 23.92 & 37.81 & 46.15 & 63.89 & 22.99 & 39.44 & 44.84 \\
        \bottomrule
    \end{tabular}
    }
    \caption{Evaluation results (\%) on \realCounterMemoryQA. 
    $\mathbb{P}(\mathbf{U_i})$ denotes the proportion of instances for which the model incorrectly update its answer.
    $\mathbb{P}(\mathbf{U_c})$ denotes the proportion of instances for which the model correctly update its answer.
    Here \texttt{Explicit Prompt} refers to the explicit prompting strategy of Assumption-in-Question. 
    \texttt{Mistral} refers to \texttt{Mistral-v0.3-7B}. 
    The baseline model is provided with 3-shot exemplars for ICL while \Ours is in zero-shot inference.
    }
    \label{tab:eval_RealCounterMemoryQA}
    \vspace{-12pt}
\end{table*}

%% file: Contents/tables/tab_real_counter_memory_QA_data_statistics.tex
\begin{table}[!ht]
    \small 
    \centering
    \setlength{\tabcolsep}{2pt}
    \scalebox{0.9}{
    \begin{tabular}{lccccc}
        \toprule
        \multirow{3}{*}{Dataset} & \multirow{3}{*}{Full Size} & \multicolumn{4}{c}{Counter-Memory Subset} \\
        \cmidrule(l){3-6}
                                 &                            &  \multicolumn{2}{c}{Mistral w/ Alpaca} & \multicolumn{2}{c}{Mistral w/ \Ours} \\
        \cmidrule(l){3-4} \cmidrule(l){5-6}
                                 &                            &  Size & Ratio (\%) & Size & Ratio (\%) \\
        \midrule
        BioASQ & 1,504 & 661 & 43.95 & 610 & 40.56 \\
        DROP & 1,503 & 1,043 & 69.39 & 1,019 & 67.80 \\
        DuoRC & 1,501 & 1,350 & 89.94 & 1,350 & 89.94 \\
        RACE & 674 & 573 & 85.01 & 569 & 84.42 \\
        RE & 2,948 & 1,892 & 64.18 & 1,787 & 60.62 \\
        TextbookQA & 1,503 & 648 & 43.11 & 611 & 40.65 \\
        \bottomrule
    \end{tabular}
    }
    \caption{
    Statistics of data subsets of \realCounterMemoryQA. 
    \texttt{Full Size} denotes the number of instances before parametric answer probing.
    \texttt{Counter-Memory} denotes the cases where the model gives a wrong parametric answer. 
    \texttt{Mistral} refers to \texttt{Mistral-v0.3-7B}.
    Results in Table~\ref{tab:eval_RealCounterMemoryQA} are based on Counter-Memory subset.
    }
    \label{tab:eval_RealCounterMemoryQA_statistics}
    \vspace{-12pt}
\end{table}

%% file: Contents/tables/tab_ablation.tex
\begin{table*}[!ht]
    \small 
    \centering
    \scalebox{1}{
    \begin{tabular}{lcccccc}
        \toprule 
        \multirow{3}{*}{Model} & \multicolumn{4}{c}{IfQA} & \multicolumn{2}{c}{MRQA} \\ 
        \cmidrule(l){2-5} \cmidrule(l){6-7}
                               & \multicolumn{2}{c}{w/ Gold Passages} & \multicolumn{2}{c}{w/ Mixed Passages} & \multicolumn{2}{c}{w/ Gold Passages} \\
        \cmidrule(l){2-3} \cmidrule(l){4-5} \cmidrule(l){6-7} 
                               & F$_1$ & EM & F$_1$ & EM  & F$_1$ & EM \\ 
        \midrule
        \Ours & 80.53 & 74.14 & \textbf{77.85} & \textbf{70.86} & \textbf{73.67} & \textbf{62.91} \\
        \quad - Random Noise Contexts & 77.76 & 71.57 & 68.99 & 62.00 & 70.67 & 61.16 \\
        \quad + Answer Derivation (before answer) & 78.40 & 70.43 & 72.52 & 64.00 & 68.06 & 57.42 \\
        \quad + Answer Derivation (after answer) & 77.76 & 71.57 & 68.99 & 62.00 & 71.93 & 62.40 \\
        \quad - Shuffling Gold Contexts \& Assumptions & \textbf{80.55} & \textbf{75.00} & 77.22 & 70.43 & 72.66 & 62.74 \\
        \bottomrule
    \end{tabular}
    }
    \caption{Ablation results (\%) on IfQA full split test set and MRQA test set. Zero-shot performance with the normal prompt is presented.}
    \label{tab:ablation}
    \vspace{-12pt}
\end{table*}

%% file: Contents/results_and_analysis.tex
\section{Results and Analysis}
\subsection{Main Results}\label{sec::Results and Analysis::Main Results}

\paragraph{Performance on IfQA}

Based on Table~\ref{tab:main_eval_ifqa_full_split} and Table~\ref{tab:eval_ifqa_full_split}, instruction-tuned LLMs generally achieve better performance than base LLMs. 
GPT-4o gives the best performance and the best robustness. 
\Ours is better than all the open-weight LLMs and is comparable to GPT-3.5 5-shot in the gold passage setting while surpassing it in the mixed passage setting.
Another observation is that all the baselines except GPT-4o are vulnerable to noise in the context passages while \Ours is much more robust.

Meanwhile, the benefit of an explicit prompting method for knowledge preference in gold passage setting is not significant.
Explicit prompting tends to be more useful when there is little noise. 
In the mixed passage setting, using explicit prompting leads to a slightly degraded performance which could be related to the noise from the retrieved passages.
This reveals that, in addition to the ability of prioritizing the target knowledge, the ability of identifying relevant knowledge is also vital to LLMs.

\paragraph{Performance on \mquakeAdapted}

According to Table~\ref{tab:main_eval_knowledge_preference_3-shot} and Table~\ref{tab:eval_knowledge_preference_3-shot}, one observation is that in 3-shot setting with explicit prompting, GPT-4o achieves the best performance in terms of both the absolute value and the ratios of the QA performance.
Then is Llama-3-8B-Instruct and \Ours which achieve similar performance.  
Meanwhile, without explicit prompting, \Ours dominates, which means inherently \Ours is better at following the hierarchical knowledge preference. 

Besides, we find that LLMs with better instruction following ability are more likely to be better in \mquakeAdapted (see our additional evaluation results on IFEval~\cite{zhou2023instruction} in Appendix~\ref{sec::Full Evaluation Results::Evaluation on IFEval}).
Llama-3-8B-Instruct and GPT-4o serve representative cases for this. 
However, the performance is not always aligned. 
For example, Mistral-v0.3-7B-Instruct is much better at instruction following but worse at \mquakeAdapted than Llama-2-7B-Instruct.
Another observation is that the gap between the top performing LLMs and other testee LLMs in \mquakeAdapted is large which further justifies that typical instruction tuning can not always improve the knowledge preference following ability.
The gap within the top performing LLMs, however, is not so huge. 
This indicates the \mquakeAdapted is not hard in terms of its requirements on the multi-hop reasoning and reading comprehension, but \mquakeAdapted essentially requires testee LLMs to follow the knowledge preference hierarchy.

Note that GPT-4o shows generally solid knowledge preference compared to all of the other baselines including GPT-3.5.
This justifies our motivation to introduce a type of instruction tuning data for modeling the hierarchical knowledge preference and also justifies our approach on synthesizing part of the instances through GPT-4o.

\paragraph{Performance on \realCounterMemoryQA and MRQA}

Table~\ref{tab:eval_mrqa_test_w_oracle_contexts} shows that \Ours largely enhances the LLM's capability in seeking and leveraging the context knowledge across different domains. 
Table~\ref{tab:eval_RealCounterMemoryQA_statistics} includes the knowledge probing results which reveal that \Ours has nearly no difference with the baseline when no context is given. 
When the context knowledge conflicts with the parametric knowledge, \Ours outperforms the baseline in terms of correcting the wrong parametric answer based on the context knowledge (see Table~\ref{tab:eval_RealCounterMemoryQA}). 
This indicates that \Ours well prioritizes the context knowledge regardless of whether the explicit prompting is adopted or not.

\subsection{Analysis of Counterfactual Single-Hop QA Data}\label{sec::Analysis of Counterfactual Single-Hop QA Data}
Fig.~\ref{fig:cf_sh_qa_data_quality} shows the test results of LLM trained with IfQA train set, our synthesized single-hop counterfactual QA data, and with a combination of them.
The test performance of the LLM tuned on the train set of the IfQA saturates, which shows that the human annotations lead to limited patterns. 
Furthermore, our synthesized data together with the train set of IfQA further improve the test set performance.
We can also see that simply tuning the LLM with our synthesized data which is generated through zero-shot prompting still cannot match the in-domain human annotated IfQA train set.

\subsection{Ablation Study}

We provide the zero-shot results of \Ours with different training strategies on IfQA and MRQA, which are both human annotated, to justify our final choice:
(i) add randomly sampled noise context passages, 
(ii) do not add the step-by-step answer derivation in training,
and (iii) randomly shuffle the oracle passages and assumptions (if possible). 
Table~\ref{tab:ablation} justifies our design choice. 
Table~\ref{tab:eval_knowledge_preference_shuffled_contexts_no-icl} and Table~\ref{tab:eval_knowledge_preference_shuffled_contexts_3-shot} further shows that shuffling the assumptions and oracle contexts can avoid LLMs to take shortcuts for multi-hop QA.

%% file: Contents/related_work.tex
\section{Related Work}

\subsection{Knowledge Conflicts}
Previous related studies have focused on the preference of language models between external context knowledge and the internal parametric knowledge~\cite{longpre-etal-2021-entity, xie2023adaptive, kortukov2024studying, qaknowledge2024}. 
\citeauthor{xie2023adaptive} finds that LLMs generally prefer evidence consistent with their parametric knowledge over the conflicting evidence~\citeyearpar{xie2023adaptive}. 
Another finding is that LLMs demonstrate strong \textit{confirmation bias} when external evidence contains consistent information with parametric knowledge which is also supported by a more recent study~\cite{kortukov2024studying}.
On the other hand, external evidences that are coherent, convincing, though conflicting with parametric knowledge can still make LLMs highly receptive to them~\citep{xie2023adaptive, kortukov2024studying}.
Different from these works, we further refine the knowledge conflicts into instruction knowledge, context knowledge, and parametric knowledge for study and we resort to regularizing LLMs' behaviors under different knowledge conflicts. 

\subsection{Improving LLMs Under Conflicts}
Existing works have investigated how to regularize the behaviors of LLMs in conflicts. 
One typical scenario is to edit new knowledge into LLM artifacts to inject external knowledge to override the parametric knowledge. 
Corresponding methods include revising the LLM weights, applying adaptor networks, and integrating explicit memories~\cite{meng2022locating, meng2022mass, de2021editing, mitchell2022memory, zhong2023mquake}.
Our work introduces the instruction knowledge to integrate the goal of this research direction with the more complex scenario where external contexts cause additional knowledge conflicts. 
Furthermore our work resort to instruction tuning to enable such knowledge injections against knowledge conflicts inherently in inference time.

Recent works have also explored improving the safety of LLMs against jailbreak attacks inside instructions. 
OpenAI has introduced the instruction hierarchy~\cite{wallace2024instruction} to teach LLMs to ignore jailbreak instructions for this direction.
In contrast, our work focuses more on the knowledge conflicts and building the preference hierarchy between the instruction as a whole, the context passages, and LLMs' parameters.

\subsection{Learning From Human Preference}
Some LLM related works have focused on aligning LLMs with human preference to direct LLMs towards being more helpful, less toxic, and more powerful~\cite{korbak2023pretraining, dai2023safe, tian2023fine, havrilla2024teaching}.
Reinforcement learning from human feedback (RLHF) is the most widely adopted technique for such alignment, with its classical pipeline comprising supervised fine-tuning, reward model training, and reinforcement learning via policy optimization~\cite{ouyang2022training}.
Various works have been developed for improved efficiency, less bias, and enhanced alignment~\cite{lu2022quark, dong2023raft, selfimprovement2024a, sun2024principle, rafailov2024direct, meng2024simpo}.
Our work has similar goals with these works to build LLMs that are more aligned to human preference.
However, different from this line of research, our work focuses more on the knowledge utilization aspect of LLMs, namely the way LLMs leverage external knowledge and parametric knowledge while adhering to human preference (instruction knowledge).

%% file: Contents/conclusion.tex
\section{Conclusion}
In this work, we unify different settings where LLMs should integrate external knowledge (e.g., user specifications, retrieved passages, and updated knowledge) with their internal knowledge by introducing instruction knowledge, context knowledge, and parametric knowledge.
We further defined a knowledge preference hierarchy over three types of knowledge as a blueprint to achieve this unified target. 
For systematic evaluation on the LLMs' knowledge preference, we compiled a collection of existing benchmarks covering different preference settings.
To teach LLMs to inherently follow this knowledge preference hierarchy, we synthesized various instruction tuning data (\Ours) with source data from Wikipedia and Wikidata.
Comprehensive evaluation and analysis show the superior performance of \Ours over vanilla instruction tuning in terms of following the knowledge preference hierarchy.
As future work, the question of \emph{how many samples will be enough for LLMs to achieve perfect knowledge preference} can be further investigated.

%% file: Contents/limitations.tex
\section*{Limitations} 
First of all, prioritizing the instruction knowledge or the knowledge provided by users leads to a fine-tuned LLMs well following the human instructions or human provided knowledge. 
Similar to related instruction tuning works, this may raise safety concerns since user instruction can also contain jailbreak attacks. 
Since the robustness of LLMs against such jailbreak attacks is not the main focus of this work, we leave this for research works on LLM safety. 
Potential solutions include further refining the instruction knowledge into system level instruction knowledge (more prioritized constraints or knowledge handled by LLM providers and customers can not modify them in applications) and user level instruction knowledge so that safety issues can be addressed. 
Another potential solution is to add a safety guard. 
Second, our prompting format of synthesized QA instances for instruction tuning can be more diverse as we currently mainly use Alpaca~\cite{alpaca}'s prompt template and surrounds different instance components with fixed tags.  
To achieve our goal of this paper, this may not be an issue.
But for real world applications, some augmentation methods might be needed to accommodate different users' prompting styles.

%% file: Contents/ethics_statements.tex
\section*{Ethics Statements}

The synthesis process is based on GPT models. 
The source data of our synthesis process may contain outdated information or facts and the synthesis process is based on GPT models. 
Hence, follow-up works adopting our synthesized data should be aware of this and further verification might be needed.
Meanwhile, we have introduced different kinds of counterfactual QA instances.
Downstream applications based on our synthesized data or corresponding instruction tuned LLMs should also be aware of this.

%% file: Contents/acknowledgements.tex
\section*{Acknowledgements}

Research was supported in part by the Institute for Geospatial Understanding through an Integrative Discovery Environment (I-GUIDE) by NSF under Award No. 2118329, US DARPA KAIROS Program No. FA8750-19-2-1004 and INCAS Program No. HR001121C0165, National Science Foundation IIS-19-56151, and the Molecule Maker Lab Institute: An AI Research Institutes program supported by NSF under Award No. 2019897. Any opinions, findings, and conclusions or recommendations expressed herein are those of the authors and do not necessarily represent the views, either expressed or implied, of DARPA or the U.S. Government.
The views and conclusions contained in this paper are those of the authors and should not be interpreted as representing any funding agencies.

%% file: Contents/appendix.tex
\input{Contents/figs/fig_counterfact_single_hop_quality}

\section{Prompt Templates}
\label{sec::Appendix::Prompt Templates}

\subsection{Alpaca Prompt Templates}
We put the prompt template used by Alpaca~\cite{alpaca} in Table~\ref{tab:alpaca_w_input_template} and Table~\ref{tab:alpaca_wo_input_template} for reference purpose.

\begingroup
\begin{table*}[ht]
    \centering
    \small
    \begin{tabular}{p{\linewidth}}
        \toprule
        \underline{\textbf{\textsc{Alpaca w/ Input}}} \\
        \vspace{-2mm}
        Below is an instruction that describes a task, paired with an input that provides further context. Write a response that appropriately completes the request.\\
        \\
        \#\#\# Instruction:\\
        \hl{\{instruction\}} \\
        \\
        \#\#\# Input:\\
        \hl{\{input\}} \\
        \\
        \#\#\# Response: \\
        \bottomrule
    \end{tabular}
    \caption{
        Alpaca prompt template with input.
        Contents which are instance specific and to be filled in are highlighted in light blue.
    }
    \label{tab:alpaca_w_input_template}
\end{table*}
\endgroup

\begingroup
\begin{table*}[ht]
    \centering
    \small
    \begin{tabular}{p{\linewidth}}
        \toprule
        \underline{\textbf{\textsc{Alpaca w/o Input}}} \\
        \vspace{-2mm}
        Below is an instruction that describes a task. Write a response that appropriately completes the request.\\
        \\
        \#\#\# Instruction:\\
        \hl{\{instruction\}} \\
        \\
        \#\#\# Response: \\
        \bottomrule
    \end{tabular}
    \caption{
        Alpaca prompt template without input.
        Contents which are instance specific and to be filled in are highlighted in light blue.
    }
    \label{tab:alpaca_wo_input_template}
\end{table*}
\endgroup

\subsection{Context-Augmented QA Prompt Template}
Table~\ref{tab:rag_qa_template} contains the prompt template based on Alpaca's prompt template for context-augmented QA. 

\begingroup
\begin{table*}[ht]
    \centering
    \small
    \begin{tabular}{p{\linewidth}}
        \toprule
        \underline{\textbf{\textsc{Context-Augmented QA Template}}} \\
        \vspace{-2mm}
        Below is an instruction that describes a task, paired with an input that provides further context. Write a response that appropriately completes the request.\\
        \\
        \#\#\# Instruction:\\
        Answer the **question** using the **retrieved documents** as reference information. Your answer should be short (a few words or an entity). Output your final **answer** enclosed by <answer> and <\/answer> tags. \\
        \\
        \hl{\{ICL Exemplars in Alpaca's \texttt{\#\#\# Input} \& \texttt{\#\#\# Response} Format if any\}} \\
        \\
        \#\#\# Input:\\
        <question> \hl{\{question\}} </question> \\
        <retrieved> \hl{\{context passages\}} </retrieved> \\
        \\
        \#\#\# Response: \\
        \bottomrule
    \end{tabular}
    \caption{
        Context-augmented QA prompt template.
        Contents which are instance specific and to be filled in are highlighted in light blue.
    }
    \label{tab:rag_qa_template}
\end{table*}
\endgroup

\subsection{Explicit Prompts for Hierarchical Knowledge Preference}\label{sec::Appendix::Prompt Templates::Strong Prompts for Hierarchical Knowledge Preference}
Table~\ref{tab:rag_qa_strong_assumption_in_question_template} contains the context-augmented prompt template with the prompting method named as Assumption-in-Question. 
It means we explicitly instruct LLMs to follow the target knowledge preference hierarchy. 
In some tasks, the instruction knowledge such as the user specifications or question assumptions can not be easily separated from the problem or the question.
So this prompt template treats the instruction knowledge is within the input and the explicit prompting method is designed to accommodate this position variation.  

Table~\ref{tab:rag_qa_strong_assumption_in_instruction_template} contains the context-augmented prompt template with the prompting method  named as Assumption-in-Instruction.
Similarly, we also explicitly instruct LLMs to follow the target knowledge preference hierarchy.
Its difference from Assumption-in-Question lies in the fact that Assumption-in-Instruction is designed for instances where the instruction knowledge can be well separated from the question or problem input.
For such instances, the assumptions will be put in the instruction section of the Alpaca's prompt, separated from the problem input as well as the context passages.

\begingroup
\begin{table*}[ht]
    \centering
    \small
    \begin{tabular}{p{\linewidth}}
        \toprule
        \underline{\textbf{\textsc{Context-Augmented QA Template w/ Assumption-in-Question Prompting}}} \\
        \vspace{-2mm}
        Below is an instruction that describes a task, paired with an input that provides further context. Write a response that appropriately completes the request.\\
        \\
        \#\#\# Instruction:\\
        Answer the **question** using the **retrieved documents** as reference information. Your answer should be short (a few words or an entity). Output your final **answer** enclosed by <answer> and <\/answer> tags. 
        \hlred{For ANY knowledge conflicts and ANY information conflicts, STRICTLY PRIORITIZE assumptions in the input question over retrieved documents, and STRICTLY PRIORITIZE the retrieved documents over your parametric knowledge.} \\
        \\
        \hl{\{ICL Exemplars in Alpaca's \texttt{\#\#\# Input} \& \texttt{\#\#\# Response} Format if any\}} \\
        \\
        \#\#\# Input:\\
        <question> \hl{\{question w/ assumption (instruction knowledge) if any\}} </question> \\
        <retrieved> \hl{\{context passages\}} </retrieved> \\
        \\
        \#\#\# Response: \\
        \bottomrule
    \end{tabular}
    \caption{
        Context-augmented QA prompt template with explicit prompting method of Assumption-in-Question. 
        Contents which are instance specific and to be filled in are highlighted in light blue.
        The injected prompt for modeling hierarchical knowledge preference is highlighted in light red. 
    }
    \label{tab:rag_qa_strong_assumption_in_question_template}
\end{table*}
\endgroup

\begingroup
\begin{table*}[ht]
    \centering
    \small
    \begin{tabular}{p{\linewidth}}
        \toprule
        \underline{\textbf{\textsc{Context-Augmented QA Template w/ Assumption-in-Instruction Prompting}}} \\
        \vspace{-2mm}
        Below is an instruction that describes a task, paired with an input that provides further context. Write a response that appropriately completes the request.
        \hlred{For ANY knowledge conflicts and ANY information conflicts, STRICTLY PRIORITIZE instruction over input and STRICTLY PRIORITIZE input over your parametric knowledge.}\\
        \\
        \#\#\# Instruction:\\
        \hl{\{assumption (instruction knowledge)\}}
        Answer the **question** using the **retrieved documents** as reference information. Your answer should be short (a few words or an entity). Output your final **answer** enclosed by <answer> and <\/answer> tags. \\
        \\
        \hl{\{ICL Exemplars in Alpaca's \texttt{Assumption} \& \texttt{\#\#\# Input} \& \texttt{\#\#\# Response} Format if any\}} \\
        \\
        Again, \hl{\{assumption (instruction knowledge)\}} \\
        \\
        \#\#\# Input:\\
        <question> \hl{\{question\}} </question> \\
        <retrieved> \hl{\{context passages\}} </retrieved> \\
        \\
        \#\#\# Response: \\
        \bottomrule
    \end{tabular}
    \caption{
        Context-augmented QA prompt template with explicit prompting method of Assumption-in-Instruction.
        Contents which are instance specific and to be filled in are highlighted in light blue.
        The injected prompt for modeling hierarchical knowledge preference is highlighted in light red. 
    }
    \label{tab:rag_qa_strong_assumption_in_instruction_template}
\end{table*}
\endgroup

\subsection{Data Synthesis Prompt Templates}\label{sec::Appendix::Prompt Templates::Data Synthesis Prompt Templates}
For the synthesis of multi-hop QA instances, the question synthesis prompt template is shown by Table~\ref{tab:mh_question_synthesis_prompt_template}.
The passage synthesis prompt template is shown by Table~\ref{tab:mh_passage_synthesis_prompt_template}.
The answer derivation prompt template is shown by Table~\ref{tab:mh_answer_derivation_synthesis_prompt_template}.

\begingroup
\begin{table*}[ht]
    \centering
    \small
    \begin{tabular}{p{\linewidth}}
        \toprule
        \underline{\textbf{\textsc{Question Synthesis for Multi-Hop QA}}} \\
        \vspace{-2mm}
        You are a powerful multi-hop question generator. Using the provided fact chain (relation triples in order), generate a multi-hop question that incorporates only the head entity (\hl{\{head entity of fact chain\}}) and all the relations from the relation triples. The tail entity (\hl{\{tail entity of fact chain\}}) should serve as the answer based on the knowledge contained within the fact chain. Ensure that the generated question excludes all entities from the fact chain, except for the head entity (\hl{\{head entity of fact chain\}}). Each relation triple should be treated as a fact.\\
        \bottomrule
    \end{tabular}
    \caption{
        Question synthesis prompt template for multi-hop QA instances (both factual or counterfactual).
        Contents which are instance specific and to be filled in are highlighted in light blue.
    }
    \label{tab:mh_question_synthesis_prompt_template}
\end{table*}
\endgroup

\begingroup
\begin{table*}[ht]
    \centering
    \small
    \begin{tabular}{p{\linewidth}}
        \toprule
        \underline{\textbf{\textsc{Passage Synthesis for Multi-Hop QA}}} \\
        \vspace{-2mm}
        Generate a realistic passage of about 50 words that supports the fact expressed by the following relation triple:\\
        <relation triple> \hl{\{relation triple\}} </relation triple> \\
        Your generated passage should avoid mentioning any other facts or details that imply different tail entities for the same head entity (\hl{\{head entity of the relation triple\}}) and relation (\hl{\{tail entity of the relation triple\}}) of the above relation triple. Meanwhile, your generated passage should avoid mentioning and also avoid conflicting with the facts expressed by all the following relation triples:\\
        \hl{\{other relation triples for synthesizing passages for this instance\}} \\
        Now, follow the above requirements and provide your generated passage enclosed by <passage> and </passage> tags.\\
        \bottomrule
    \end{tabular}
    \caption{
        Passage synthesis prompt template for multi-hop QA instances (both factual or counterfactual).
        Contents which are instance specific and to be filled in are highlighted in light blue.
    }
    \label{tab:mh_passage_synthesis_prompt_template}
\end{table*}
\endgroup

\begingroup
\begin{table*}[ht]
    \centering
    \small
    \begin{tabular}{p{\linewidth}}
        \toprule
        \underline{\textbf{\textsc{Answer Derivation for Multi-Hop QA}}} \\
        \vspace{-2mm}
        Given the multi-hop question, the answer, and the relation triples as the underlying gold knowledge required to derive the answer, generate a coherent, concise, and step-by-step explanation for how to derive the answer based on the question and the knowledge contained within the relation triples.\\
        While you should leverage the information encapsulated in the relation triples, avoid explicitly mentioning the triples themselves. Instead, focus on presenting each piece of knowledge as if the knowledge was summarized from some reference documents.\\
        <question> \hl{\{synthesized question\}} </question> \\
        <answer> \hl{\{answer\}} </answer>\\
        <gold knowledge>  \hl{\{relation triples from the fact chain\}} </gold knowledge> \\
        Now, provide your generated answer explanation enclosed by <explanation> and </explanation> tags. \\
        \bottomrule
    \end{tabular}
    \caption{
        Answer derivation synthesis prompt template for multi-hop QA instances (both factual or counterfactual).
        Contents which are instance specific and to be filled in are highlighted in light blue.
    }
    \label{tab:mh_answer_derivation_synthesis_prompt_template}
\end{table*}
\endgroup

For the synthesis of counterfactual single-hop QA instances, the prompt template is shown by Table~\ref{tab:sh_cf_question_answer_derivation_synthesis_prompt_template}.
For the synthesis of factual single-hop QA instances, the prompt template is shown by Table~\ref{tab:sh_f_question_answer_derivation_synthesis_prompt_template}.

\begingroup
\begin{table*}[ht]
    \centering
    \small
    \begin{tabular}{p{\linewidth}}
        \toprule
        \underline{\textbf{\textsc{Question, Answer, and Answer Derivation Synthesis for Single-Hop Counterfactual QA}}} \\
        \vspace{-2mm}
        Based on the provided passage and your knowledge, generate a challenging counterfactual question answer pair and the corresponding concise and step-by-step answer derivation explanation. The question must introduce counterfactual and hypothetical conditions or incidents. The answer must:\\
        1. be PRECISE (avoid vagueness, uncertainty, and vague quantifiers such as 'fewer', 'less', 'longer', 'increased', etc.),\\
        2. be CONCISE (an entity or a few words),\\
        3. be CHALLENGING to get (avoid simple negation of facts or other trivial answers), and\\
        4. be UNIQUELY DERIVABLE with counterfactual reasoning based on the passage, the hypothetical question, and commonsense. If the provided passage lacks sufficient information (e.g., external knowledge or specific commonsense is needed) to make sure the answer is uniquely derivable, further provide the additional information as an additional realistic passage enclosed by <passage> and </passage> tags.\\
        \\
        The generated question should be enclosed by <question> and </question> tags, the generated answer should be enclosed by <answer> and </answer> tags, and the generated answer derivation explanation should be enclosed by <explanation> and </explanation> tags.\\
        Here is the provided passage:\\
        <passage> \hl{\{Wikipedia passage\}} </passage> \\
        \bottomrule
    \end{tabular}
    \caption{
        Question, answer, and answer derivation synthesis prompt template for single-hop counterfactual QA instances.
        Contents which are instance specific and to be filled in are highlighted in light blue.
    }
    \label{tab:sh_cf_question_answer_derivation_synthesis_prompt_template}
\end{table*}
\endgroup

\begingroup
\begin{table*}[ht]
    \centering
    \small
    \begin{tabular}{p{\linewidth}}
        \toprule
        \underline{\textbf{\textsc{Question, Answer, and Answer Derivation Synthesis for Single-Hop Factual QA}}} \\
        \vspace{-2mm}
        Based on the provided passage and your knowledge, generate a challenging question answer pair and the corresponding concise and step-by-step answer derivation explanation. \\
        The answer must:\\
        1. be PRECISE (avoid vagueness, uncertainty, and vague quantifiers such as 'fewer', 'less', 'longer', 'increased', etc.),\\
        2. be CONCISE (an entity or a few words),\\
        3. be CHALLENGING to get (avoid trivial answers), and\\
        4. be UNIQUELY DERIVABLE with reasoning based on the passage. If the provided passage lacks sufficient information (e.g., external knowledge is needed) to make sure the answer is uniquely derivable, further provide the additional information as an additional realistic passage enclosed by <passage> and </passage> tags.\\
        \\
        The generated question should be enclosed by <question> and </question> tags, the generated answer should be enclosed by <answer> and </answer> tags, and the generated answer derivation explanation should be enclosed by <explanation> and </explanation> tags.\\
        Here is the provided passage:\\
        <passage> \hl{\{Wikipedia passage\}} </passage> \\
        \bottomrule
    \end{tabular}
    \caption{
        Question, answer, and answer derivation synthesis prompt template for single-hop factual QA instances.
        Contents which are instance specific and to be filled in are highlighted in light blue.
    }
    \label{tab:sh_f_question_answer_derivation_synthesis_prompt_template}
\end{table*}
\endgroup

\section{Implementation Details}\label{sec::Appendix::Implementation Details}

\subsection{Fact Chain Mining}\label{sec::Appendix::Fact Chain Mining}
The fact chain mining is conducted in a dense subset of Wikidata\footnote{\href{https://kopl.xlore.cn/index}{WikiData15k}} which contains 16960 entities, 794 concepts, 363 relations, and 846 properties. 
The following heuristic rules or requirements are applied\footnote{Some of the heuristic rules are adapted from MQuAKE to make sure the multi-hop question can be fluent and natural~\cite{zhong2023mquake}.}:
(1) no repeated entities or relations in the fact chain,
(2) the fact chain contains up to 3 different entity concepts,
(3) triples with a country tail entity can only appear in the last two hops,
(4) all triples with a person or location tail entity are consecutive,
(5) the head entity for a relation triple with relation \texttt{headquarters location} must be an organization entity and the head entity for a relation triple with relation \texttt{capital} must be a country entity,
(6) for original fact chain mining, given the head entity and the relation, the tail entity must be unique within the subgraph,
(7) for fact chain editing, the newly factually updated tail entity should be unique within the subgraph given the head entity and relation (otherwise the fact chain editing will be abandoned),
(7) max number of child nodes for exploration in the BFS search is set to 5,
(8) the edited or the factually updated tail entity and the original tail entity are of the same concept,
and (9) avoid including entities which are concepts.

For converting fact chain edits to counterfactual assumptions, we adopt a fixed template. 
Namely, given a list of original triples to be edited and a list of corresponding edited triples, we have the counterfactual assumption as: ``\textit{Assume the following relation triples hold true: [List of original relation triples], and assume the following relation triples do not hold true any more: [List of corresponding edited relation triples].}''.

\subsection{Data Synthesis}\label{sec::Appendix::Statistics of Synthesized Data}
For parametric answer probing, we using the similar prompt template in Table~\ref{tab:rag_qa_template} and we heuristically consider the parametric answer as identical to gold-standard answer if the F$_1$ score exceeds 0.80 if there is not an exact match.

For calling GPT-4o, we set temperature as 0.6 for multi-hop QA instances and 0.9 for single-hop QA instances. 
The max\_tokens is set to 4096 while the top\_p is set to 1.
Fig.~\ref{fig:synthesized_data_statistics} shows the distribution of \Ours synthesized data.

\input{Contents/figs/fig_synthesized_data_statistics}

\subsection{Instruction Tuning}
To augment the synthesized data for instruction tuning, we randomly sample 2 different passages and 3 different passages for single-hop and multi-hop instances respectively as noise passages.
The noise passages are placed before the randomly shuffled context passages as we expect that, with a qualified retriever, irrelevant passages should be easily identified and put closer to the middle of the LLMs' input~\cite{liu2024lost}. 
For counterfactual multi-hop QA instances whose assumptions can be separated from the question, we also randomly sample the assumptions to avoid LLMs to take shortcuts in training.

We fine-tune our main LLMs based on LoRA~\cite{hu2021lora} (target modules: q\_proj, k\_proj, v\_proj, o\_proj, and rank: 16), with batch size as 128, learning rate as 1e-4 (searched from \{5e-5, 1e-4, 3e-4\}), max length as 2048, warmup steps as 100, number of epochs as 10, saving and evaluation periods as 200 steps. 
We randomly sample 2000 instances as the validation set and pick the checkpoint with lowest validation loss for evaluations.

For analytical experiments in Sec.~\ref{sec::Analysis of Counterfactual Single-Hop QA Data} focusing on IfQA, we use the same hyperparameters except that we set the learning rate as 3e-4 (as it achieves better performance), the number of epochs as 15, warmup steps as 0, saving and evaluation frequency as per epoch.
The best performance among all checkpoints is reported as these analytical experiments aim to comparing the performance upper bounds.

\subsection{Evaluation}
For evaluation on different test sets, we adopt the official evaluation script of MRQA~\cite{fisch2019mrqa} for normalizing the answers and calculating F$_1$ and EM metrics.

For inference with LLMs in this paper, we generally use the temperature as 0.6 and the top\_p as 0.9.
For \mquakeAdapted, max new tokens for generation is set as 256. 
For IfQA, max new tokens for generation is set to 256. 
For MRQA, max new tokens for generation is set to 128.

\section{Full Evaluation Results}
Due to the limited space for main contents, we put the complete experimental results here.

\subsection{Statistics of Evaluation Data}
Table~\ref{tab:eval_datasets_statistics} shows the brief statistics about the major evaluation datasets on which we have conducted our experiments.  
\input{Contents/tables/tab_eval_data_statistics}

\subsection{Evaluation on IfQA}
\input{Contents/tables/tab_ifqa_full-split}

Table~\ref{tab:eval_ifqa_full_split} shows the evaluation results on the test set of IfQA~\cite{yu-etal-2023-ifqa} full split.

\subsection{Evaluation on \mquakeAdapted}

\input{Contents/tables/tab_knowledge_preference_3-shot}

\input{Contents/tables/tab_knowledge_preference_no-icl}

\input{Contents/tables/tab_knowledge_preference_context_shuffled_no-icl}

\input{Contents/tables/tab_knowledge_preference_context_shuffled_3-shot}

Table~\ref{tab:eval_knowledge_preference_3-shot} contains the evaluation results on \mquakeAdapted with 3-shot in-context learning.
Table~\ref{tab:eval_knowledge_preference_no-icl} contains the evaluation results on \mquakeAdapted with zero-shot. 
Since \mquakeAdapted contains multi-hop QA instances, to avoid providing shortcuts through presenting LLMs with context passages in the same order as the relation triples in the fact chain, we shuffle context passages, leading to \mquakeAdapted With Shuffled Contexts, and conduct the same evaluations.
The corresponding zero-shot and 3-shot evaluation results on \mquakeAdapted With Shuffled Contexts are shown in Table~\ref{tab:eval_knowledge_preference_shuffled_contexts_no-icl} and Table~\ref{tab:eval_knowledge_preference_shuffled_contexts_3-shot} respectively.

\subsection{Evaluation on IFEval}\label{sec::Full Evaluation Results::Evaluation on IFEval}
\input{Contents/tables/tab_ifeval}

To investigate the correlation between LLMs' instruction following ability and the knowledge preference following ability, we evaluate four LLMs (Mistral-v0.3-7B w/ Alpaca, Mistral-v0.3-7B w/ \Ours, GPT-3.5, and gpt-4o) on IFEval~\cite{zhou2023instruction}.
To adapt Alpaca's prompt template for base LLMs, we set the contents of the instruction section as ``\textit{Strictly follow the request in the input.}'' and the contents of the input section as the target prompts.
Other parts of the setup are the same as the Open LLM Leaderboard v2~\cite{open-llm-leaderboard-v2}.
The results together with baseline scores from Open LLM Leaderboard v2~\cite{open-llm-leaderboard-v2} and original paper~\cite{zhou2023instruction} are shown in Table~\ref{tab:ifeval_accuracy-summarization}. 
We find that the instruction following ability and the knowledge preference ability correlate but are not perfectly aligned (see analysis in Sec.~\ref{sec::Results and Analysis::Main Results}).

\section{Case Study}
\begin{figure*}[!ht]
    \centering
    \includegraphics[width=\linewidth]{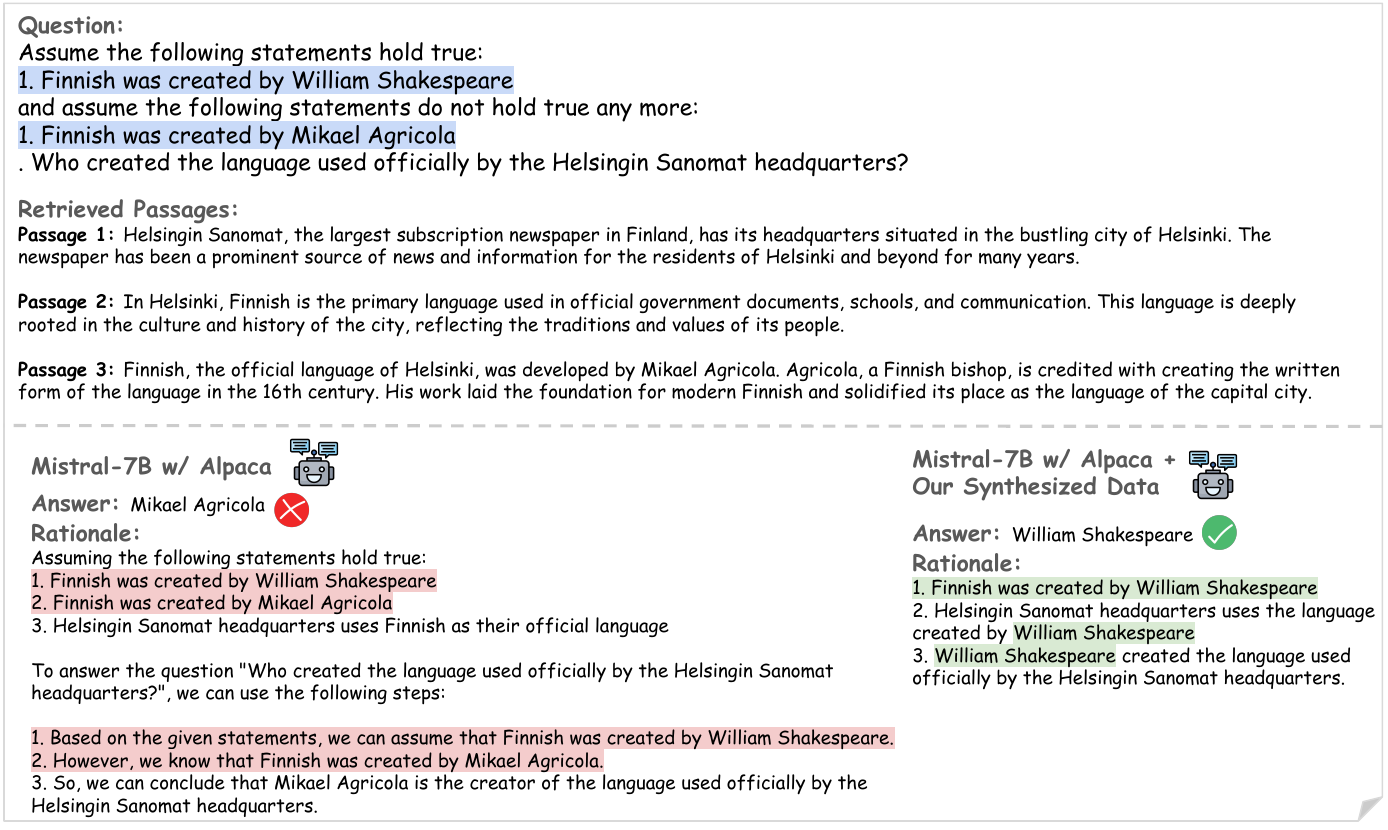}
    \caption{Case study for LLMs' preference for the instruction knowledge. In the figure, the instruction knowledge is highlighted in light blue. }
    \label{fig:case_study_instruction_knowledge}
    \vspace{-12pt}
\end{figure*}

\begin{figure*}[!ht]
    \centering
    \includegraphics[width=\linewidth]{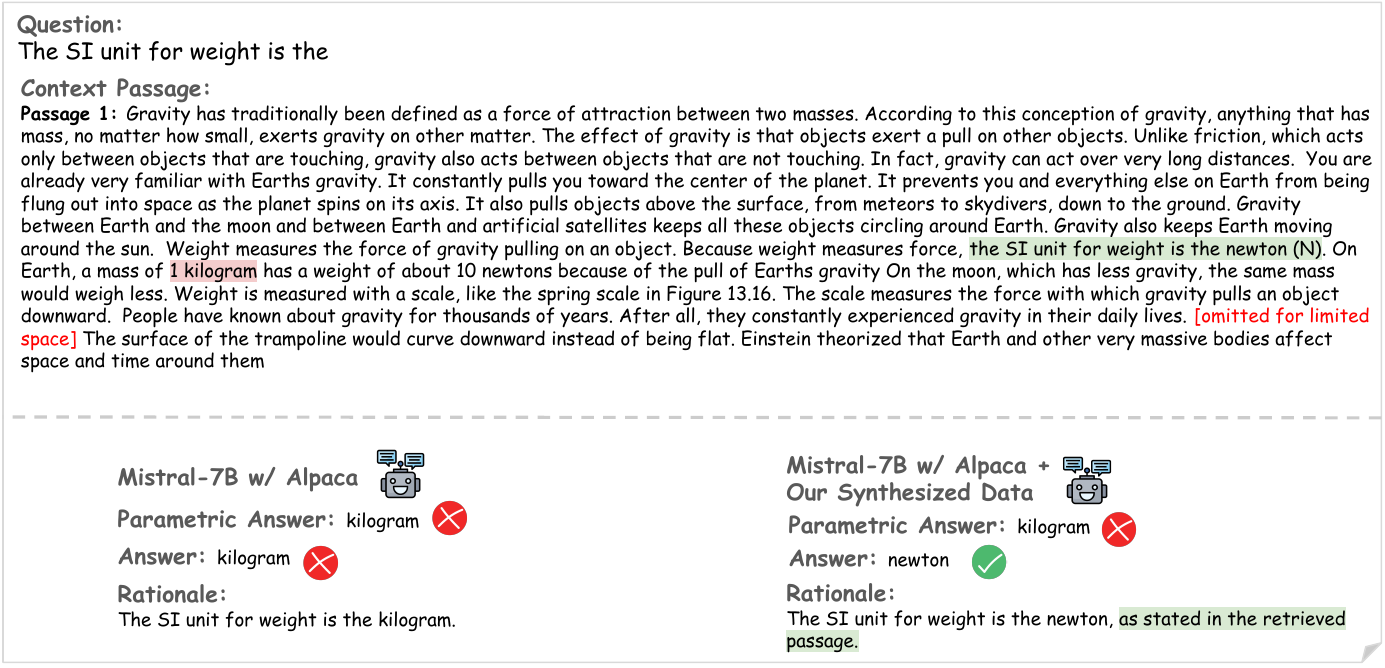}
    \caption{Case study for LLMs' preference for context knowledge. }
    \label{fig:case_study_context_knowledge}
    \vspace{-12pt}
\end{figure*}

To complement quantitative studies, we also conduct case studies as shown in Fig.~\ref{fig:case_study_instruction_knowledge} and Fig.~\ref{fig:case_study_context_knowledge}.
The corresponding baseline LLM conducts inference with explicit prompts and with 3-shot in-context exemplars while our model is in zero-shot inference setting.
To obtain the answer derivation rationale, we concatenate the input and output of corresponding models and further append \texttt{<derivation>} to continue the generation.

Fig.~\ref{fig:case_study_instruction_knowledge} shows that both LLMs well capture the instruction knowledge and the context knowledge. 
The difference is that the baseline LLM with conventional instruction tuning still prefers the context knowledge over the instruction knowledge in conflicting scenario. 
In contrast, \Ours coherently and consistently prioritizes and integrates the instruction knowledge with its reasoning over the context knowledge, leading to the correct answer.

Fig.~\ref{fig:case_study_context_knowledge} shows that both the baseline LLM and \Ours have the wrong parametric answer. 
However, even given the context passage, the baseline LLM still sticks to its own parametric knowledge while \Ours prioritizes the context passages to derive the correct answer. 
This indicates the effectiveness of \Ours in terms of prioritizing the context knowledge over the parametric knowledge.

%% file: Contents/figs/fig_counterfact_single_hop_quality.tex
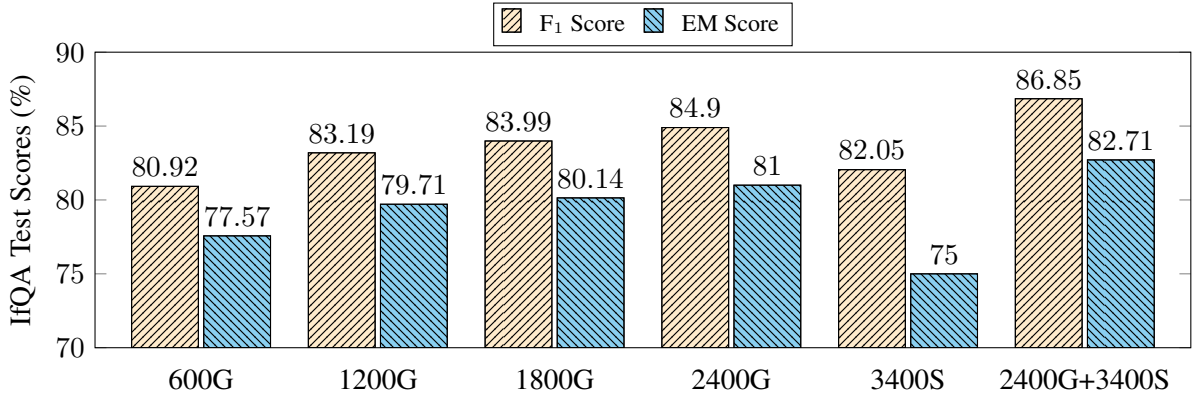
\begin{figure*}[htb]
    \begin{center}
    {
        \pgfplotsset{compat=1.13,
            /pgfplots/ybar legend/.style={
            /pgfplots/legend image code/.code={%
               \draw[##1,/tikz/.cd,yshift=-0.25em]
                (0cm,0cm) rectangle (7pt,0.8em);},
           },
        }
        \pgfplotsset{width=16cm, height=5.5cm}
            \centering
            
            \begin{tikzpicture}  
                \begin{axis}  
                [  
                    ybar,
                    ymin=70, ymax=90,
                    ytick={70,75,80,85,90,95},
                    major x tick style = transparent,
                    bar width=25pt,
                    enlarge x limits=0.12,
                    ylabel={IfQA Test Scores (\%)},
                    symbolic x coords={0,1,2,3,4,5},  
                    xtick=data,  
                    xticklabels={600G, 1200G, 1800G, 2400G, 3400S, 2400G+3400S},
                    nodes near coords,  
                    nodes near coords align={vertical},  
                    legend cell align=left,
                    legend columns=2 row=1,
                    legend style={
                        at={(0.5,1.02)},
                        anchor=south,
                        column sep=1ex,
                        font=\small,
                    }
                ]  
                \addplot[ybar, fill=blanchedalmond,  postaction={pattern=north east lines}] coordinates {
                    (0, 80.92)
                    (1, 83.19)
                    (2, 83.99)
                    (3, 84.90)
                    (4, 82.05)
                    (5, 86.85)
                };  
                \addplot[ybar, fill=babyblue,  postaction={pattern=north west lines}] coordinates {
                    (0, 77.57)
                    (1, 79.71)
                    (2, 80.14)
                    (3, 81.00)
                    (4, 75.00)
                    (5, 82.71)
                };
                \legend{F$_1$ Score, EM Score} 
                \end{axis}  
            \end{tikzpicture}
            \caption{Evaluation scores on IfQA test set of the full split. Note that \texttt{G} denotes that the training data is from IfQA's train set while \texttt{S} denotes that the training data is from \Ours synthesized single-hop QA set. The number before \texttt{G} or \texttt{S} represents the corresponding size of data used. \label{fig:cf_sh_qa_data_quality}}
    }
    \end{center}
\end{figure*}

%% file: Contents/figs/fig_synthesized_data_statistics.tex
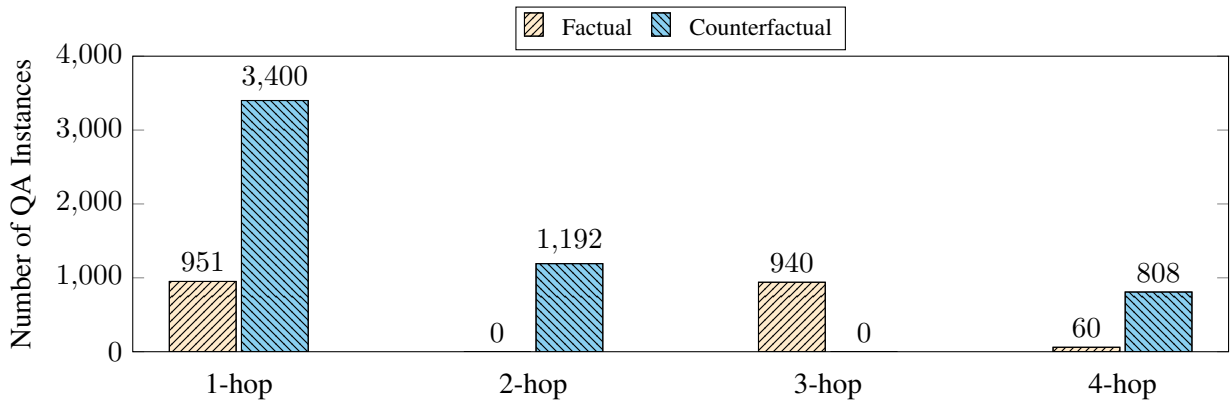
\begin{figure*}[htb]
    \begin{center}
    {
        \pgfplotsset{compat=1.13,
            /pgfplots/ybar legend/.style={
            /pgfplots/legend image code/.code={%
               \draw[##1,/tikz/.cd,yshift=-0.25em]
                (0cm,0cm) rectangle (7pt,0.8em);},
           },
        }
        \pgfplotsset{width=16cm, height=5.5cm}
            \centering
            
            \begin{tikzpicture}  
                \begin{axis}  
                [  
                    ybar,
                    ymin=0, ymax=4000,
                    ytick={0,1000,2000,3000,4000},
                    major x tick style = transparent,
                    bar width=25pt,
                    enlarge x limits=0.12,
                    ylabel={Number of QA Instances},
                    symbolic x coords={0,1,2,3},  
                    xtick=data,  
                    xticklabels={1-hop, 2-hop, 3-hop, 4-hop},
                    nodes near coords,  
                    nodes near coords align={vertical},  
                    legend cell align=left,
                    legend columns=2 row=1,
                    legend style={
                        at={(0.5,1.02)},
                        anchor=south,
                        column sep=1ex,
                        font=\small,
                    }
                ]  
                \addplot[ybar, fill=blanchedalmond,  postaction={pattern=north east lines}] coordinates {
                    (0, 951)
                    (1, 0)
                    (2, 940)
                    (3, 60)
                };  
                \addplot[ybar, fill=babyblue,  postaction={pattern=north west lines}] coordinates {
                    (0, 3400)
                    (1, 1192)
                    (2, 0)
                    (3, 808)
                };
                \legend{Factual, Counterfactual} 
                \end{axis}  
            \end{tikzpicture}
            \caption{Statistics of \Ours synthesized data. \label{fig:synthesized_data_statistics}}
    }
    \end{center}
\end{figure*}

%% file: Contents/tables/tab_eval_data_statistics.tex
\begin{table}[!ht]
    \small 
    \centering
    \begin{tabular}{lccc}
        \toprule
        Dataset & \#QA instances \\
        \midrule
        IfQA & 700 \\
        \mquakeAdapted & 9,000\\
        MRQA & 9,633 \\
        \bottomrule
    \end{tabular}
    \caption{Brief statistics about datasets evaluated.}
    \label{tab:eval_datasets_statistics}
\end{table}

%% file: Contents/tables/tab_ifqa_full-split.tex
\begingroup
\begin{table*}[!ht]
    \small 
    \centering
    \renewcommand\tabcolsep{4pt} 
    \scalebox{1}{
    \begin{tabular}{lccccccccc}
        \toprule 
        \multirow{3}{*}{Model} & \multirow{3}{*}{\# Shots} &  \multicolumn{4}{c}{Normal Prompt} & \multicolumn{4}{c}{Explicit Prompt} \\ 
                               &                              & \multicolumn{2}{c}{w/ Gold Passages} & \multicolumn{2}{c}{w/ Mixed Passages} & \multicolumn{2}{c}{w/ Gold Passages} & \multicolumn{2}{c}{w/ Mixed Passages} \\
        \cmidrule(l){3-4} \cmidrule(l){5-6} \cmidrule(l){7-8} \cmidrule(l){9-10}
                               &                              & F$_1$ & EM & F$_1$ & EM  & F$_1$ & EM & F$_1$ & EM \\ 
        \midrule
        \rowcolor[gray]{0.85}
        \multicolumn{10}{c}{\textit{\textbf{Closed-Source LLMs}}}   \\ 
        \midrule
        \multirow{3}{*}{GPT-3.5 Turbo} & 0 & 74.83 & 66.42 & 61.53 & 51.00 & 70.71 & 60.71 & 57.17 & 47.29 \\
                                       & 3 & 76.59 & 70.29 & 71.55 & 65.86 & 76.94 & 71.57 & 71.06 & 66.00 \\
                                       & 5 & 77.70 & 71.86 & 73.27 & 67.57 & 79.70 & 74.14 & 72.24 & 66.57 \\
        \multirow{3}{*}{GPT-4o} & 0 & 88.09 & 80.43 & 85.39 & 77.86 & 88.19 & 80.71 & 85.38 & 77.29 \\
                                & 3 & 89.56 & 83.29 & 87.12 & 80.71 & 90.18 & 84.43 & 87.87 & 81.29 \\
                                & 5 & 90.43 & 84.57 & 87.50 & 81.14 & 89.71 & 83.86 & 87.88 & 81.57 \\ 
        \midrule
        \rowcolor[gray]{0.85}
        \multicolumn{10}{c}{\textit{\textbf{Open-Weight LLMs}}}   \\ 
        \midrule
        \multirow{3}{*}{Llama-2-7B} & 0 & 26.42 & 17.86 & 14.86 & 7.43 & 27.19 & 18.00 & 13.67 & 7.29 \\
                                    & 3 & 40.06 & 32.00 & 24.86 & 18.71 & 39.63 & 31.43 & 24.63 & 18.71 \\
                                    & 5 & 35.96 & 29.29 & 22.27 & 16.29 & 35.85 & 28.57 & 20.21 & 14.14 \\
        \multirow{3}{*}{Llama-2-7B-Instruct} & 0 & 30.72 & 21.29 & 13.94 & 5.71 & 29.01 & 20.29 & 11.81 & 4.57 \\
                                             & 3 & 52.47 & 43.43 & 30.54 & 22.29 & 51.26 & 43.00 & 28.77 & 21.14 \\
                                             & 5 & 40.12 & 30.71 & 9.11 & 4.43 & 40.19 & 30.57 & 7.40 & 3.57 \\
        \multirow{3}{*}{Mistral-v0.3-7B} & 0 & 49.98 & 42.14 & 35.01 & 27.71 & 45.64 & 37.57 & 31.66 & 25.14 \\
                                    & 3 & 59.52 & 52.14 & 42.34 & 36.43 & 59.56 & 53.43 & 40.27 & 35.00 \\
                                    & 5 & 57.94 & 51.57 & 35.38 & 29.71 & 56.11 & 50.14 & 34.09 & 28.57 \\
        \multirow{3}{*}{Mistral-v0.3-7B-Instruct} & 0 & 46.32 & 30.57 & 36.52 & 24.43 & 44.95 & 29.43 & 33.16 & 22.29 \\
                                             & 3 & 67.38 & 58.14 & 58.69 & 49.71 & 68.79 & 59.00 & 57.63 & 48.14 \\
                                             & 5 & 71.26 & 63.14 & 59.13 & 51.71 & 70.76 & 62.29 & 57.03 & 49.71 \\
        \multirow{3}{*}{Qwen-2-7B} & 0 & 49.41 & 41.00 & 22.26 & 14.57 & 46.28 & 37.43 & 26.72 & 20.29 \\
                                 & 3 & 65.20 & 58.29 & 43.60 & 36.86 & 63.29 & 56.71 & 41.62 & 35.14 \\
                                 & 5 & 65.56 & 58.57 & 41.00 & 35.43 & 65.03 & 58.43 & 39.06 & 33.14 \\
        \multirow{3}{*}{Qwen-2-7B-Instruct} & 0 & 64.76 & 58.14 & 44.04 & 36.71 & 63.99 & 56.29 & 45.08 & 37.71 \\
                                          & 3 & 70.67 & 63.57 & 50.79 & 44.00 & 70.92 & 63.29 & 51.27 & 44.00 \\
                                          & 5 & 70.04 & 62.43 & 50.96 & 43.29 & 70.64 & 62.71 & 48.28 & 41.43 \\ 
        \multirow{3}{*}{Llama-3-8B} & 0 & 48.25 & 40.71 & 31.66 & 25.57 & 47.90 & 41.29 & 29.91 & 23.71 \\
                                    & 3 & 54.99 & 49.14 & 42.81 & 37.14 & 55.95 & 50.29 & 42.82 & 36.00 \\
                                    & 5 & 58.47 & 52.29 & 42.24 & 36.43 & 56.91 & 50.57 & 44.57 & 38.14 \\
        \multirow{3}{*}{Llama-3-8B-Instruct} & 0 & 70.30 & 62.00 & 49.63 & 43.57 & 67.27 & 59.71 & 48.27 & 41.43 \\
                                             & 3 & 71.60 & 65.00 & 58.29 & 50.43 & 71.44 & 64.57 & 59.03 & 51.57 \\
                                             & 5 & 74.50 & 68.86 & 60.09 & 53.00 & 75.33 & 69.14 & 58.00 & 51.00 \\
        \midrule
        \rowcolor[gray]{0.85}
        \multicolumn{10}{c}{\textit{\textbf{Ours}}}   \\ 
        \midrule
        \multirow{3}{*}{Mistral-v0.3-7B w/ Alpaca} & 0 & 54.16 & 45.14 & 31.15 & 22.29 & 52.51 & 44.29 & 28.54 & 20.71 \\
                                                   & 3 & 68.05 & 61.43 & 46.47 & 40.29 & 68.38 & 61.43 & 47.78 & 40.29 \\
                                                   & 5 & 67.98 & 61.71 & 50.71 & 44.00 & 67.22 & 60.29 & 49.49 & 43.14 \\                                        
        \multirow{1}{*}{Mistral-v0.3-7B w/  \Ours} & 0 & 80.53 & 74.14 & 77.85 & 70.86 & 80.53 & 73.86 & 77.33 & 70.29 \\
        \bottomrule
    \end{tabular}
    }
    \caption{All evaluation results on IfQA full split test set. Assumption-in-Question is adopted for \texttt{Explicit Prompt}.}
    \label{tab:eval_ifqa_full_split}
\end{table*}
\endgroup

%% file: Contents/tables/tab_knowledge_preference_3-shot.tex
\begin{table*}[!ht]
    \small 
    \centering
    \scalebox{0.8}{
    \begin{tabular}{lccccccccc}
        \toprule 
        \multirow{2}{*}{Model} & \multirow{2}{*}{Gold Ans.} & \multicolumn{2}{c}{2-hop} & \multicolumn{2}{c}{3-hop} & \multicolumn{2}{c}{4-hop} & \multicolumn{2}{c}{Overall} \\ 
        \cmidrule(l){3-4} \cmidrule(l){5-6} \cmidrule(l){7-8} \cmidrule(l){9-10}
                               &                            & F$_1$ & EM & F$_1$ & EM & F$_1$ & EM & F$_1$ & EM \\ 
        \midrule
        \rowcolor[gray]{0.85}
        \multicolumn{10}{c}{\textit{\textbf{Explicit Prompt: Assumption-in-Instruction}}}   \\ 
        \midrule
        \multirow{2}{*}{GPT-3.5 Turbo} & Ori. & 51.24 & 48.43 & 43.81 & 41.03 & 46.77 & 42.70 & 47.27 & 44.06 \\
                                       & New  & 42.88 & 41.03 & 49.70 & 47.47 & 44.34 & 43.00 & 45.64 & 43.83 \\
        \multirow{2}{*}{GPT-4o} & Ori. & 2.86 & 0.17 & 3.64 & 2.00 & 2.60 & 1.43 & 3.03 & 1.20 \\
                               & New  & 94.44 & 93.83 & 93.40 & 92.50 & 97.01 & 96.13 & 94.95 & 94.16 \\
        \multirow{2}{*}{Llama-2-7B} & Ori. & 71.35 & 68.53 & 47.21 & 44.00 & 54.64 & 53.17 & 57.73 & 55.23 \\
                                    & New  & 23.80 & 20.90 & 45.56 & 44.00 & 39.05 & 38.70 & 36.14 & 34.53 \\
        \multirow{2}{*}{Llama-2-7B-Instruct} & Ori. & 24.45 & 21.30 & 19.82 & 17.37 & 23.99 & 22.20 & 22.76 & 20.29 \\
                                             & New  & 60.88 & 59.17 & 66.52 & 65.57 & 61.91 & 61.37 & 63.10 & 62.03 \\
        \multirow{2}{*}{Llama-3-8B} & Ori. & 44.43 & 41.20 & 47.14 & 45.20 & 45.96 & 44.57 & 45.84 & 43.66  \\
                                    & New  & 49.54 & 47.67 & 44.73 & 43.20 & 45.13 & 44.57 & 46.47 & 45.14 \\
        \multirow{2}{*}{Llama-3-8B-Instruct} & Ori. & 5.53 & 2.73 & 5.70 & 3.97 & 12.79 & 11.50 & 8.01 & 6.07 \\
                                             & New  & 92.86 & 92.10 & 90.84 & 89.90 & 85.37 & 84.20 & 89.69 & 88.73 \\
        \multirow{2}{*}{Qwen-2-7B} & Ori. & 34.41 & 32.20 & 29.26 & 27.87 & 33.12 & 31.87 & 32.26 & 30.64 \\
                                    & New  & 60.87 & 59.53 & 64.05 & 63.10 & 63.58 & 63.13 & 62.83 & 61.92 \\
        \multirow{2}{*}{Qwen-2-7B-Instruct} & Ori. & 12.22 & 9.53 & 24.17 & 22.67 & 26.17 & 24.90 & 20.85 & 19.03 \\
                                             & New  & 81.78 & 80.87 & 63.03 & 61.63 & 55.21 & 54.23 & 66.67 & 65.58 \\
        \multirow{2}{*}{Mistral-v0.3-7B} & Ori. & 50.24 & 47.00 & 35.24 & 33.30 & 40.20 & 38.97 & 41.89 & 39.76 \\
                                         & New  & 44.90 & 43.10 & 59.40 & 57.63 & 55.24 & 54.46 & 53.18 & 51.73 \\
        \multirow{2}{*}{Mistral-v0.3-7B-Instruct} & Ori. & 40.57 & 37.10 & 34.29 & 31.87 & 44.64 & 39.13 & 39.84 & 36.03 \\
                                                  & New  & 44.02 & 41.57 & 50.71 & 48.50 & 43.18 & 42.03 & 45.97 & 44.03 \\
        \multirow{2}{*}{Mistral-v0.3-7B w/ Alpaca} & Ori. & 74.00 & 71.83 & 65.77 & 64.17 & 71.82 & 69.83 & 70.53 & 68.61 \\
                                                   & New  & 22.18 & 19.33 & 28.66 & 26.03 & 22.60 & 21.70 & 24.48 & 22.36 \\
        \multirow{2}{*}{Mistral-v0.3-7B w/ \Ours} & Ori. & 6.32 & 3.33 & 10.81 & 9.20 & 13.91 & 12.43 & 10.35 & 8.32 \\
                                                  & New  & 92.63 & 92.07 & 86.01 & 85.10 & 84.45 & 82.90 & 87.70 & 86.69 \\
        \midrule
        \rowcolor[gray]{0.85}
        \multicolumn{10}{c}{\textit{\textbf{Explicit Prompt: Assumption-in-Question}}}   \\ 
        \midrule
        \multirow{2}{*}{GPT-3.5 Turbo} & Ori. & 62.19 & 59.57 & 48.61 & 44.73 & 52.83 & 47.97 & 54.54 & 50.76 \\
                                       & New  & 31.40 & 29.03 & 42.10 & 39.87 & 33.14 & 31.83 & 35.55 & 33.58 \\
        \multirow{2}{*}{GPT-4o} & Ori. & 3.75 & 1.10 & 5.41 & 3.77 & 5.40 & 4.20 & 4.86 & 3.02 \\
                                & New  & 94.34 & 93.63 & 91.39 & 90.40 & 94.40 & 93.60 & 93.37 & 92.54 \\
        \multirow{2}{*}{Llama-2-7B} & Ori. & 50.85 & 47.43 & 40.52 & 36.57 & 35.85 & 34.40 & 42.41 & 39.47 \\
                                    & New  & 33.91 & 31.47 & 43.62 & 41.77 & 54.07 & 53.93 & 43.87 & 42.39 \\
        \multirow{2}{*}{Llama-2-7B-Instruct} & Ori. & 43.81 & 40.20 & 29.74 & 28.13 & 15.09 & 13.10 & 29.55 & 27.14 \\
                                             & New  & 26.23 & 23.37 & 20.67 & 19.40 & 17.62 & 17.17 & 21.51 & 19.98 \\
        \multirow{2}{*}{Llama-3-8B} & Ori. & 51.55 & 48.50 & 46.92 & 44.83 & 40.49 & 39.20 & 46.32 & 44.18 \\
                                    & New  & 38.38 & 36.10 & 42.86 & 41.27 & 45.65 & 45.33 & 42.30 & 40.90 \\
        \multirow{2}{*}{Llama-3-8B-Instruct} & Ori. & 57.02 & 54.07 & 39.68 & 36.77 & 42.23 & 39.57 & 46.31 & 43.47 \\
                                             & New  & 18.78 & 15.57 & 34.75 & 32.87 & 25.83 & 25.33 & 26.45 & 24.59 \\
        \multirow{2}{*}{Qwen-2-7B} & Ori. & 52.33 & 50.03 & 47.65 & 46.03 & 43.88 & 42.63 & 47.95 & 46.23 \\
                                    & New  & 38.92 & 37.03 & 41.89 & 40.37 & 47.25 & 46.43 & 42.69 & 41.28 \\
        \multirow{2}{*}{Qwen-2-7B-Instruct} & Ori. & 54.32 & 51.80 & 55.41 & 53.57 & 53.05 & 50.93 & 54.26 & 52.10 \\
                                             & New  & 27.01 & 24.60 & 25.90 & 23.71 & 25.10 & 24.37 & 26..00 & 24.23 \\                                             
        \multirow{2}{*}{Mistral-v0.3-7B} & Ori. & 47.27 & 43.70 & 34.88 & 32.30 & 36.83 & 35.63 & 39.66 & 37.21 \\
                                         & New  & 39.96 & 37.77 & 53.50 & 51.53 & 53.40 & 52.77 & 48.95 & 47.36 \\
        \multirow{2}{*}{Mistral-v0.3-7B-Instruct} & Ori. & 47.07 & 42.63 & 39.19 & 35.37 & 44.76 & 36.80 & 43.67 & 38.27 \\
                                                  & New  & 27.45 & 24.43 & 37.11 & 34.50 & 35.71 & 34.43 & 33.42 & 31.12 \\
        \multirow{2}{*}{Mistral-v0.3-7B w/ Alpaca} & Ori. & 59.74 & 56.40 & 50.90 & 47.73 & 60.71 & 58.63 & 57.12 & 54.26 \\
                                                   & New  & 25.15 & 21.93 & 34.11 & 31.77 & 26.17 & 25.33 & 28.48 & 26.34 \\
        \multirow{2}{*}{Mistral-v0.3-7B w/ \Ours} & Ori. & 4.97 & 1.97 & 6.95 & 5.17 & 12.16 & 10.87 & 8.03 & 6.00 \\
                                                  & New  & 92.97 & 92.40 & 90.14 & 89.27 & 85.36 & 83.40 & 89.49 & 88.36 \\
        \midrule
        \rowcolor[gray]{0.85}
        \multicolumn{10}{c}{\textit{\textbf{Normal Prompt: Alpaca}}}   \\ 
        \midrule
        \multirow{2}{*}{GPT-3.5 Turbo} & Ori. & 64.79 & 61.63 & 49.29 & 45.37 & 53.62 & 48.73 & 55.90 & 51.91 \\
                                       & New  & 28.48 & 26.17 & 41.44 & 39.23 & 32.33 & 31.07 & 34.08 & 32.16 \\
        \multirow{2}{*}{GPT-4o} & Ori. & 5.56 & 3.00 & 12.44 & 10.87 & 16.00 & 14.70 & 11.33 & 9.52 \\
                                & New  & 92.11 & 91.17 & 83.61 & 82.63 & 83.64 & 83.03 & 86.46 & 85.61 \\
        \multirow{2}{*}{Llama-2-7B} & Ori. & 49.04 & 45.23 & 39.70 & 35.93 & 36.96 & 35.67 & 41.90 & 38.94 \\
                                    & New  & 35.78 & 33.67 & 43.98 & 42.27 & 53.59 & 53.40 & 44.44 & 43.11 \\
        \multirow{2}{*}{Llama-2-7B-Instruct} & Ori. & 43.54 & 39.97 & 32.90 & 30.80 & 19.64 & 17.10 & 32.03 & 29.29 \\
                                             & New  & 29.40 & 26.40 & 25.73 & 24.43 & 23.00 & 22.63 & 26.04 & 24.49 \\
        \multirow{2}{*}{Llama-3-8B} & Ori. & 51.32 & 48.50 & 45.30 & 43.30 & 40.28 & 39.00 & 45.64 & 43.60 \\
                                    & New  & 39.51 & 37.40 & 44.06 & 42.27 & 45.57 & 45.23 & 43.05 & 41.63 \\
        \multirow{2}{*}{Llama-3-8B-Instruct} & Ori. & 58.72 & 55.93 & 40.03 & 37.13 & 43.21 & 40.67 & 47.32 & 44.58 \\
                                             & New  & 18.95 & 15.80 & 34.55 & 32.60 & 25.71 & 25.20 & 26.41 & 24.53 \\
        \multirow{2}{*}{Qwen-2-7B} & Ori. & 49.18 & 46.93 & 46.12 & 44.30 & 44.26 & 43.00 & 46.52 & 44.74 \\
                                    & New  & 41.53 & 40.20 & 43.10 & 41.50 & 47.59 & 46.83 & 44.07 & 42.84 \\
        \multirow{2}{*}{Qwen-2-7B-Instruct} & Ori. & 56.93 & 54.67 & 58.23 & 56.60 & 56.70 & 54.53 & 57.29 & 55.27 \\
                                             & New  & 25.81 & 23..23 & 26.45 & 24.37 & 25.08 & 24.43 & 25.78 & 23.98 \\
        \multirow{2}{*}{Mistral-v0.3-7B} & Ori. & 47.95 & 44.03 & 35.76 & 33.03 & 36.95 & 35.87 & 40.22 & 37.64 \\
                                         & New  & 39.44 & 37.50 & 52.05 & 50.03 & 52.99 & 52.40 & 48.16 & 46.64 \\
        \multirow{2}{*}{Mistral-v0.3-7B-Instruct} & Ori. & 47.20 & 42.63 & 39.23 & 35.30 & 44.86 & 36.80 & 43.76 & 38.24 \\
                                                  & New  & 28.07 & 25.20 & 36.71 & 34.20 & 35.24 & 33.97 & 33.34 & 31.12 \\
        \multirow{2}{*}{Mistral-v0.3-7B w/ Alpaca} & Ori. & 60.16 & 56.83 & 49.52 & 46.30 & 59.71 & 57.87 & 56.46 & 53.67 \\
                                                   & New  & 24.86 & 21.77 & 34.76 & 32.23 & 25.60 & 24.83 & 28.40 & 26.28 \\
        \multirow{2}{*}{Mistral-v0.3-7B w/ \Ours} & Ori. & 5.08 & 2.07 & 6.79 & 5.13 & 12.82 & 11.37 & 8.23 & 6.19 \\
                                                  & New  & 93.25 & 92.73 & 90.02 & 89.13 & 84.81 & 82.87 & 89.36 & 88.24 \\
        \bottomrule
    \end{tabular}
    }
    \caption{3-shot evaluation results on \mquakeAdapted.}
    \label{tab:eval_knowledge_preference_3-shot}
\end{table*}

%% file: Contents/tables/tab_knowledge_preference_no-icl.tex
\begin{table*}[!ht]
    \small 
    \centering
    \scalebox{0.9}{
    \begin{tabular}{lccccccccc}
        \toprule 
        \multirow{2}{*}{Model} & \multirow{2}{*}{Gold Ans.} & \multicolumn{2}{c}{2-hop} & \multicolumn{2}{c}{3-hop} & \multicolumn{2}{c}{4-hop} & \multicolumn{2}{c}{Overall} \\ 
        \cmidrule(l){3-4} \cmidrule(l){5-6} \cmidrule(l){7-8} \cmidrule(l){9-10}
                               &                            & F$_1$ & EM & F$_1$ & EM & F$_1$ & EM & F$_1$ & EM \\ 
        \midrule
        \rowcolor[gray]{0.85}
        \multicolumn{10}{c}{\textit{\textbf{Explicit Prompt: Assumption-in-Instruction}}}   \\ 
        \midrule
        \multirow{2}{*}{Llama-2-7B} & Ori. & 37.53 & 31.67 & 29.00 & 24.47 & 42.18 & 39.13 & 36.24 & 31.76 \\
                                    & New  & 30.74 & 27.47 & 23.44 & 20.30 & 21.09 & 19.20 & 25.09 & 22.32 \\
        \multirow{2}{*}{Llama-2-7B-Instruct} & Ori. & 6.75 & 3.77 & 13.30 & 11.07 & 11.96 & 9.57 & 10.67 & 8.13 \\
                                             & New  & 80.28 & 78.28 & 63.52 & 60.23 & 65.75 & 62.77 & 69.85 & 67.07 \\
        \multirow{2}{*}{Llama-3-8B} & Ori. & 28.91 & 26.07 & 36.98 & 34.50 & 50.07 & 48.33 & 38.65 & 36.30 \\
                                    & New  & 59.93 & 58.53 & 45.31 & 43.50 & 37.28 & 36.43 & 47.51 & 46.16 \\
        \multirow{2}{*}{Llama-3-8B-Instruct} & Ori. & 10.83 & 7.77 & 8.06 & 6.10 & 11.40 & 10.17 & 10.10 & 8.01 \\
                                             & New  & 86.93 & 85.87 & 88.39 & 86.97 & 87.10 & 86.13 & 87.48 & 86.32 \\
        \multirow{2}{*}{Qwen-2-7B} & Ori. & 20.58 & 17.87 & 21.65 & 19.83 & 29.95 & 28.93 & 24.06 & 22.21 \\
                                    & New  & 66.31 & 65.07 & 63.89 & 62.47 & 62.08 & 60.07 & 64.09 & 62.53 \\
        \multirow{2}{*}{Qwen-2-7B-Instruct} & Ori. & 9.27 & 6.77 & 11.03 & 9.47 & 20.10 & 18.90 & 13.47 & 11.71 \\
                                             & New  & 82.46 & 81.33 & 75.30 & 73.67 & 68.82 & 66.97 & 72.52 & 73.99 \\                                             
        \multirow{2}{*}{Mistral-v0.3-7B} & Ori. & 30.28 & 26.30 & 32.55 & 29.90 & 43.18 & 40.93 & 35.34 & 32.37 \\
                                         & New  & 62.42 & 60.77 & 55.43 & 53.03 & 43.78 & 42.53 & 53.88 & 52.11 \\
        \multirow{2}{*}{Mistral-v0.3-7B-Instruct} & Ori. & 28.11 & 22.83 & 35.62 & 31.63 & 50.97 & 46.30 & 38.23 & 33.59 \\
                                                  & New  & 51.17 & 43.50 & 41.81 & 34.10 & 36.95 & 33.17 & 43.31 & 36.92 \\
        \multirow{2}{*}{Mistral-v0.3-7B w/ Alpaca} & Ori. & 66.43 & 61.47 & 63.43 & 59.17 & 74.43 & 70.60 & 68.10 & 63.74 \\
                                                   & New  & 20.20 & 16.53 & 21.83 & 18.30 & 11.85 & 10.47 & 17.96 & 15.10 \\
        \multirow{2}{*}{Mistral-v0.3-7B w/ \Ours} & Ori. & 2.99 & 0.07 & 2.57 & 0.90 & 7.15 & 6.07 & 4.23 & 2.34 \\
                                                  & New  & 96.23 & 95.77 & 95.62 & 94.70 & 92.63 & 91.53 & 94.83 & 94.00 \\
        \midrule
        \rowcolor[gray]{0.85}
        \multicolumn{10}{c}{\textit{\textbf{Explicit Prompt: Assumption-in-Question}}}   \\ 
        \midrule
        \multirow{2}{*}{Llama-2-7B} & Ori. & 26.84 & 15.90 & 21.38 & 11.97 & 35.28 & 24.20 & 27.83 & 17.36 \\
                                    & New  & 18.45 & 12.67 & 14.89 & 8.73 & 12.25 & 7.23 & 15.20 & 9.54 \\
        \multirow{2}{*}{Llama-2-7B-Instruct} & Ori. & 24.45 & 12.57 & 14.42 & 4.53 & 12.83 & 1.23 & 17.23 & 6.11 \\
                                             & New  & 14.72 & 8.20 & 10.30 & 2.23 & 8.70 & 0.77 & 11.24 & 3.73 \\
        \multirow{2}{*}{Llama-3-8B} & Ori. & 43.24 & 39.27 & 49.72 & 46.93 & 62.51 & 60.50 & 51.83 & 48.90 \\
                                    & New  & 39.15 & 40.20 & 43.13 & 41.00 & 44.38 & 43.87 & 43.47 & 41.69 \\
        \multirow{2}{*}{Llama-3-8B-Instruct} & Ori. & 45.61 & 42.70 & 46.55 & 43.53 & 46.28 & 44.53 & 46.15 & 43.59 \\
                                             & New  & 42.90 & 40.20 & 43.13 & 41.00 & 44.38 & 43..87 & 43.47 & 41.69 \\
        \multirow{2}{*}{Qwen-2-7B} & Ori. & 30.04 & 26.80 & 36.43 & 34.50 & 46.92 & 45.63 & 37.80 & 35.64 \\
                                    & New  & 45.26 & 44.00 & 40.22 & 38.50 & 42.05 & 40.37 & 42.51 & 40.96 \\
        \multirow{2}{*}{Qwen-2-7B-Instruct} & Ori. & 37.02 & 34.50 & 38.72 & 36.77 & 49.79 & 48.70 & 41.84 & 39.99 \\
                                             & New  & 33.65 & 31.37 & 26.41 & 25.07 & 16.63 & 15.77 & 25.56 & 24.07 \\                                             
        \multirow{2}{*}{Mistral-v0.3-7B} & Ori. & 26.19 & 20.17 & 31.53 & 26.37 & 40.62 & 37.27 & 32.78 & 27.93 \\
                                         & New  & 53.13 & 50.00 & 41.15 & 36.63 & 33.88 & 32.03 & 42.72 & 39.56 \\
        \multirow{2}{*}{Mistral-v0.3-7B-Instruct} & Ori. & 17.36 & 12.73 & 25.35 & 21.47 & 33.69 & 29.40 & 25.47 & 21.20 \\
                                                  & New  & 56.90 & 49.20 & 48.72 & 41.23 & 44.88 & 39.90 & 50.17 & 43.44 \\
        \multirow{2}{*}{Mistral-v0.3-7B w/ Alpaca} & Ori. & 43.41 & 36.77 & 47.97 & 42.20 & 59.81 & 56.53 & 50.30 & 45.17 \\
                                                   & New  & 36.08 & 31.77 & 28.26 & 24.17 & 21.34 & 19.80 & 28.56 & 25.24 \\
        \multirow{2}{*}{Mistral-v0.3-7B w/ \Ours} & Ori. & 2.91 & 0.00 & 2.21 & 00.53 & 6.71 & 5.60 & 3.94 & 2.04 \\
                                                  & New  & 97.12 & 96.60 & 95.92 & 94.93 & 93.14 & 92.23 & 95.39 & 94.59 \\
        \midrule
        \rowcolor[gray]{0.85}
        \multicolumn{10}{c}{\textit{\textbf{Normal Prompt: Alpaca}}}   \\ 
        \midrule
        \multirow{2}{*}{Llama-2-7B} & Ori. & 24.88 & 11.57 & 20.34 & 9.23 & 34.36 & 22.60 & 26.53 & 14.47 \\
                                    & New  & 16.12 & 7.80 & 12.79 & 5.23 & 11.73 & 6.53 & 13.55 & 6.52 \\
        \multirow{2}{*}{Llama-2-7B-Instruct} & Ori. & 20.31 & 10.87 & 19.83 & 10.87 & 14.68 & 3.40 & 18.27 & 8.38 \\
                                             & New  & 29.67 & 23.77 & 15.32 & 7.00 & 11.34 & 3.30 & 18.78 & 11.36 \\
        \multirow{2}{*}{Llama-3-8B} & Ori. & 39.49 & 35.77 & 47.08 & 44.87 & 55.43 & 54.00 & 47.33 & 44.88 \\
                                    & New  & 42.49 & 40.90 & 37.40 & 35.87 & 34.44 & 33.97 & 38.11 & 36.91 \\
        \multirow{2}{*}{Llama-3-8B-Instruct} & Ori. & 56.74 & 53.13 & 52.30 & 48.87 & 53.43 & 51.40 & 54.16 & 51.13 \\
                                             & New  & 28.29 & 25.40 & 36.39 & 34.20 & 34.88 & 34.43 & 33.19 & 31.34 \\
        \multirow{2}{*}{Qwen-2-7B} & Ori. & 36.72 & 33.90 & 41.31 & 39.50 & 48.54 & 46.93 & 42.19 & 40.11 \\
                                    & New  & 41.29 & 39.70 & 38.65 & 36.83 & 40.60 & 38.80 & 40.18 & 38.44 \\
        \multirow{2}{*}{Qwen-2-7B-Instruct} & Ori. & 36.66 & 34.17 & 41.81 & 39.77 & 48.87 & 47.70 & 42.45 & 40.54 \\
                                             & New  & 36.37 & 34.17 & 28.40 & 26.93 & 16..09 & 15.30 & 26.95 & 25.47 \\ 
        \multirow{2}{*}{Mistral-v0.3-7B} & Ori. & 27.88 & 23.50 & 33.18 & 29.90 & 40.57 & 38.13 & 33.88 & 30.51 \\
                                         & New  & 57.90 & 55.70 & 48.10 & 45.33 & 37.22 & 35.97 & 47.74 & 45.67 \\
        \multirow{2}{*}{Mistral-v0.3-7B-Instruct} & Ori. & 23.11 & 17.77 & 28.56 & 24.60 & 36.46 & 31.67 & 29.38 & 24.68 \\
                                                  & New  & 46.68 & 39.53 & 43.22 & 36.17 & 41.57 & 37.43 & 43.82 & 37.71 \\
        \multirow{2}{*}{Mistral-v0.3-7B w/ Alpaca} & Ori. & 53.06 & 47.57 & 51.00 & 46.20 & 65.71 & 62.77 & 56.59 & 52.18 \\
                                                   & New  & 30.99 & 27.43 & 30.08 & 26.17 & 20.04 & 18.87 & 27.04 & 24.16 \\
        \multirow{2}{*}{Mistral-v0.3-7B w/ \Ours} & Ori. & 2.83 & 0.00 & 2.32 & 0.63 & 7.35 & 6.23 & 4.17 & 2.29 \\
                                                  & New  & 97.22 & 96.70 & 95.89 & 94.90 & 92.45 & 91.53 & 95.19 & 94.38 \\
        \bottomrule
    \end{tabular}
    }
    \caption{Zero-shot evaluation results on \mquakeAdapted.}
    \label{tab:eval_knowledge_preference_no-icl}
\end{table*}

%% file: Contents/tables/tab_knowledge_preference_context_shuffled_no-icl.tex
\begin{table*}[!ht]
    \small 
    \centering
    \begin{tabular}{lccccccccc}
        \toprule 
        \multirow{2}{*}{Model} & \multirow{2}{*}{Gold Ans.} & \multicolumn{2}{c}{2-hop} & \multicolumn{2}{c}{3-hop} & \multicolumn{2}{c}{4-hop} & \multicolumn{2}{c}{Overall} \\ 
        \cmidrule(l){3-4} \cmidrule(l){5-6} \cmidrule(l){7-8} \cmidrule(l){9-10}
                               &                            & F$_1$ & EM & F$_1$ & EM & F$_1$ & EM & F$_1$ & EM \\ 
        \midrule
        \rowcolor[gray]{0.85}
        \multicolumn{10}{c}{\textit{\textbf{Explicit Prompt: Assumption-in-Instruction}}}   \\ 
        \midrule
        \multirow{2}{*}{Mistral-v0.3-7B w/ Alpaca} & Ori. & 63.07 & 58.77 & 53.76 & 49.93 & 59.15 & 56.00 & 58.66 & 54.90 \\
                                                   & New  & 26.44 & 22.67 & 32.30 & 28.73 & 29.80 & 28.30 & 29.51 & 26.57 \\
        \multirow{2}{*}{Mistral-v0.3-7B w/ \Ours} & Ori. & 3.31 & 0.43 & 2.17 & 0.53 & 3.05 & 2.03 & 2.85 & 1.00 \\
                                                  & New  & 95.91 & 95.40 & 95.97 & 95.00 & 96.74 & 95.67 & 96.21 & 95.36 \\
        \midrule
        \rowcolor[gray]{0.85}
        \multicolumn{10}{c}{\textit{\textbf{Explicit Prompt: Assumption-in-Question}}}   \\ 
        \midrule
        \multirow{2}{*}{Mistral-v0.3-7B w/ Alpaca} & Ori. & 45.36 & 38.90 & 44.65 & 39.43 & 48.72 & 45.80 & 46.24 & 41.38 \\
                                                   & New  & 35.52 & 31.40 & 33.39 & 29.10 & 34.46 & 33.40 & 34.46 & 31.30 \\
        \multirow{2}{*}{Mistral-v0.3-7B w/ \Ours} & Ori. & 2.88 & 0.00 & 1.95 & 0.30 & 3.38 & 2.33 & 2.73 & 0.88 \\
                                                  & New  & 97.25 & 96.70 & 96.19 & 95.20 & 96.34 & 95.10 & 96.59 & 95.67 \\
        \midrule
        \rowcolor[gray]{0.85}
        \multicolumn{10}{c}{\textit{\textbf{Normal Prompt: Alpaca}}}   \\ 
        \midrule
        \multirow{2}{*}{Mistral-v0.3-7B w/ Alpaca} & Ori. & 54.69 & 49.30 & 47.86 & 42.97 & 54.61 & 51.80 & 52.39 & 48.02 \\
                                                   & New  & 31.02 & 27.30 & 34.52 & 30.90 & 33.29 & 32.00 & 32.94 & 30.07 \\
        \multirow{2}{*}{Mistral-v0.3-7B w/ \Ours} & Ori. & 2.90 & 0.00 & 1.95 & 0.27 & 3.11 & 2.03 & 2.65 & 0.77 \\
                                                  & New  & 97.28 & 96.73 & 96.10 & 95.10 & 96.58 & 95.43 & 96.65 & 95.76 \\
        \bottomrule
    \end{tabular}
    \caption{Zero-shot evaluation results on \mquakeAdapted With Shuffled Contexts.}
    \label{tab:eval_knowledge_preference_shuffled_contexts_no-icl}
\end{table*}

%% file: Contents/tables/tab_knowledge_preference_context_shuffled_3-shot.tex
\begin{table*}[!ht]
    \small 
    \centering
    \begin{tabular}{lccccccccc}
        \toprule 
        \multirow{2}{*}{Model} & \multirow{2}{*}{Gold Ans.} & \multicolumn{2}{c}{2-hop} & \multicolumn{2}{c}{3-hop} & \multicolumn{2}{c}{4-hop} & \multicolumn{2}{c}{Overall} \\ 
        \cmidrule(l){3-4} \cmidrule(l){5-6} \cmidrule(l){7-8} \cmidrule(l){9-10}
                               &                            & F$_1$ & EM & F$_1$ & EM & F$_1$ & EM & F$_1$ & EM \\ 
        \midrule
        \rowcolor[gray]{0.85}
        \multicolumn{10}{c}{\textit{\textbf{Explicit Prompt: Assumption-in-Instruction}}}   \\ 
        \midrule
        \multirow{2}{*}{Mistral-v0.3-7B w/ Alpaca} & Ori. & 68.88 & 66.83 & 58.42 & 56.80 & 56.14 & 54.57 & 61.15 & 59.40 \\
                                                   & New  & 27.89 & 25.03 & 37.04 & 34.57 & 39.44 & 38.70 & 34.79 & 32.77 \\
        \multirow{2}{*}{Mistral-v0.3-7B w/ \Ours} & Ori. & 7.34 & 4.37 & 8.21 & 6.63 & 7.82 & 6.50 & 7.79 & 5.83 \\
                                                  & New  & 91.95 & 91.30 & 89.08 & 88.17 & 90.55 & 88.60 & 90.52 & 89.36 \\
        \midrule
        \rowcolor[gray]{0.85}
        \multicolumn{10}{c}{\textit{\textbf{Explicit Prompt: Assumption-in-Question}}}   \\ 
        \midrule
        \multirow{2}{*}{Mistral-v0.3-7B w/ Alpaca} & Ori. & 57.32 & 54.17 & 44.20 & 41.17 & 42.91 & 40.93 & 48.14 & 45.42 \\
                                                   & New  & 29.95 & 26.80 & 42.66 & 40.40 & 43.73 & 43.10 & 38.78 & 36.77 \\
        \multirow{2}{*}{Mistral-v0.3-7B w/ \Ours} & Ori. & 5.01 & 1.93 & 4.55 & 2.90 & 5.72 & 4.43 & 5.09 & 3.09 \\
                                                  & New  & 93.20 & 92.63 & 92.44 & 91.53 & 92.38 & 90.47 & 92.67 & 91.54 \\
        \midrule
        \rowcolor[gray]{0.85}
        \multicolumn{10}{c}{\textit{\textbf{Normal Prompt: Alpaca}}}   \\ 
        \midrule
        \multirow{2}{*}{Mistral-v0.3-7B w/ Alpaca} & Ori. & 57.99 & 54.53 & 45.24 & 41.97 & 42.56 & 40.33 & 48.60 & 45.61 \\
                                                   & New  & 28.99 & 25.90 & 41.33 & 38.90 & 43.42 & 42.93 & 37.91 & 35.91 \\
        \multirow{2}{*}{Mistral-v0.3-7B w/ \Ours} & Ori. & 4.79 & 1.73 & 4.36 & 2.63 & 5.99 & 4.83 & 5.04 & 3.07 \\
                                                  & New  & 93.41 & 92.80 & 92.86 & 92.00 & 92.47 & 90.63 & 92.91 & 91.81 \\
        \bottomrule
    \end{tabular}
    \caption{3-shot evaluation results on \mquakeAdapted With Shuffled Contexts.}
    \label{tab:eval_knowledge_preference_shuffled_contexts_3-shot}
\end{table*}

%% file: Contents/tables/tab_ifeval.tex
\begin{table*}[!ht]
    \small
    \centering
    \begin{tabular}{lcccc}
    \toprule
    \multirow{2}{*}{Model}  & Prompt-level  & Inst-level    & Prompt-level & Inst-level \\
                            & strict-accuracy (\%) & strict-accuracy (\%) & loose-accuracy (\%) & loose-accuracy (\%) \\
    \midrule
    GPT-4          & 76.89 & 83.57 & 79.30 & 85.37 \\
    GPT-3.5         & 63.59 & 72.90 & 65.99 & 75.42 \\
    GPT-4o          & 80.96 & 86.45 & 85.95 & 90.17 \\
    PaLM 2 S       & 43.07 & 55.76 & 46.95 & 59.11 \\
    Qwen-2-7B       & 25.32 & 37.65 & 29.02 & 41.61  \\
    Qwen-2-7B-Instruct          & 52.31 & 61.27 & 55.82 & 64.75 \\
    Llama-2-7B      & 18.48 & 31.89 & 20.89 & 34.05 \\
    Llama-2-7B-Instruct        & 32.90 & 46.40 & 44.73 & 57.19 \\
    Llama-3-8B      & 9.80 & 19.30 & 10.91 & 20.50 \\
    Llama-3-8B-Instruct        & 69.87 & 78.30 & 77.08 & 83.93 \\
    Mistral-v0.3-7B   & 15.71 & 29.62 & 16.45 & 30.70 \\
    Mistral-v0.3-7B-Instruct   & 49.35 & 59.95 & 53.05 & 63.91 \\
    Mistral-v0.3-7B w/ Alpaca & 47.13 & 58.27 & 50.28 & 61.87 \\
    Mistral-v0.3-7B w/ \Ours & 47.13 & 57.79 & 50.83 & 61.15 \\
    \bottomrule
    \end{tabular}
    \caption{Overall instruction following accuracy according to IFEval.}
    \label{tab:ifeval_accuracy-summarization}
\end{table*}